\def\year{2021}\relax
\documentclass[letterpaper]{article} % DO NOT CHANGE THIS
\usepackage{aaai21}  % DO NOT CHANGE THIS
\usepackage{times}  % DO NOT CHANGE THIS
\usepackage{helvet} % DO NOT CHANGE THIS
\usepackage{courier}  % DO NOT CHANGE THIS
\usepackage[hyphens]{url}  % DO NOT CHANGE THIS
\usepackage{graphicx} % DO NOT CHANGE THIS
\def\UrlFont{\rm}  % DO NOT CHANGE THIS
\usepackage{natbib}  % DO NOT CHANGE THIS AND DO NOT ADD ANY OPTIONS TO IT
\usepackage{caption} % DO NOT CHANGE THIS AND DO NOT ADD ANY OPTIONS TO IT
\graphicspath{{figures/}}
\usepackage{framed}

\usepackage{amsmath}
\usepackage{amssymb}
\usepackage{booktabs}
\DeclareMathOperator*{\argmax}{arg\,max}

\title{Robustness and Transferability of Universal Attacks on Compressed Models}

\author{
    Alberto G. Matachana\textsuperscript{\rm 1}, %Garc\'ia 
    Kenneth T. Co\textsuperscript{\rm 1, 2},\\
    Luis Mu\~noz-Gonz\'alez\textsuperscript{\rm 1},
    David Martinez\textsuperscript{\rm 2},
    Emil C. Lupu\textsuperscript{\rm 1}%\thanks{With help from the AAAI Publications Committee.}
    \\
}
\affiliations{
    \textsuperscript{\rm 1}Imperial College London,\hspace{0.1em}  \textsuperscript{\rm 2}DataSpartan\\
    ag4116@imperial.ac.uk, \hspace{0.1em} k.co@imperial.ac.uk
}

\begin{document}

\maketitle

\begin{abstract}
Neural network compression methods like pruning and quantization are very effective at efficiently deploying Deep Neural Networks (DNNs) on edge devices. However, DNNs remain vulnerable to adversarial examples--inconspicuous inputs that are specifically designed to fool these models. In particular, Universal Adversarial Perturbations (UAPs), are a powerful class of adversarial attacks which create adversarial perturbations that can generalize across a large set of inputs. In this work, we analyze the effect of various compression techniques to UAP attacks, including different forms of pruning and quantization. We test the robustness of compressed models to white-box and transfer attacks, comparing them with their uncompressed counterparts on CIFAR-10 and SVHN datasets. Our evaluations reveal clear differences between pruning methods, including Soft Filter and Post-training Pruning. We observe that UAP transfer attacks between pruned and full models are limited, suggesting that the systemic vulnerabilities across these models are different. This finding has practical implications as using different compression techniques can blunt the effectiveness of black-box transfer attacks. We show that, in some scenarios, quantization can produce gradient-masking, giving a false sense of security. Finally, our results suggest that conclusions about the robustness of compressed models to UAP attacks is application dependent, observing different phenomena in the two datasets used in our experiments.
\end{abstract}

\section{Introduction}
\label{sec:intro}

Fuelled by the recent advances in Deep Neural Networks (DNNs), machine learning algorithms have become the algorithm of choice for many applications such as image classification \cite{krimage}, object detection \cite{redmon2016you}, or machine translation \cite{bahdanau}, to cite but a few. Despite these impressive successes, existing DNNs face two key challenges. First, their large number of parameters causes large computational and storage overheads \cite{canziani2017analysis, han2015learning, li2017pruning}. Such overheads hinder the deployment of these algorithms on memory and power-limited devices, thus, the reduction of the complexity of these models is essential for their practical application in embedded systems, mobile, and Internet-of-Things (IoT) devices \cite{han2015learning, eie, scnn, deepiot}. Second, machine learning algorithms have been shown to be vulnerable to adversarial examples: inputs that appear similar to genuine data, but are designed to deceive the model \cite{biggio2013evasion, szegedy2014intriguing}.

Adversarial examples are a relevant threat against DNNs as these greatly undermine the usability of these models and can compromise the security and safety of systems relying on machine learning \cite{dalvi2004adversarial, lowd2005adversarial, munoz2019challenges, hau2020ghostbuster}. More worryingly, there exist Universal Adversarial Perturbations (UAPs), a class of adversarial perturbations where a single perturbation can cause a target model to misclassify on a large set of inputs \cite{moosavi2017universal}. On top of their universality, UAPs have also shown to be \textit{transferable} across models \cite{moosavi2017universal, co2019procedural}, which can enable black-box transfer attacks that do not require direct access to the victim's model \cite{papernot2016transferability, papernot2017practical, demontis2019adversarial}. Existing works have used UAPs to conduct practical attacks against image classification \cite{brown2017adversarial}, person recognition \cite{thys2019fooling}, object detection \cite{liu2018dpatch}, and to enable very query-efficient black-box attacks on DNNs \cite{co2019procedural}. These results demonstrate that UAPs are a very powerful and a realistic threat against existing DNNs.

In this paper, we provide a novel analysis on the robustness and transferability of universal attacks on compressed DNNs through the lens of UAPs. Although recent works have analysed the robustness of different compression methods that are typically used for the deployment of DNNs on edge devices against adversarial examples, these have yet to explore robustness against universal attacks in the form of UAPs. In real-world settings, adversarial attacks on models deployed in real environments need to be robust to arbitrary changes to the input. The ability for UAPs to generalize across large sets of inputs makes them a very appealing choice for the attacker and also enables physically-realizable attacks in different real-world applications. Attackers can conveniently exploit the universality and transferability of UAPs to fool DNNs deployed on small devices by using pre-computed perturbations, as there is no need for on-board computational capacity to generate UAPs. In contrast, input-specific adversarial examples need to be computed \emph{on the fly} and have to be continuously updated due to constant changes in the environment \cite{co2019universal}.

We present an empirical evaluation on the robustness and attack transferability for machine learning models compressed through pruning and quantization. We analyze whether the vulnerabilities of the original full precision model remain, change, or worsen after pruning or quantizing the model. We study the differences between the UAP robustness of the compression methods and discuss their transferability as well as vulnerability to targeted UAPs. In summary, we make the following contributions in this paper:

\begin{enumerate}
    % \item We perform white-box UAP evasion attacks on models trained with various compression mechanisms. We observed that quantization can lead to  ``gradient masking'', which could give a false sense of security, as models are still vulnerable to transfer attacks.
    \item We evaluate the transferability of UAPs across different compressed models. We observe that Soft Filter Pruning, which removes neurons or filters at training time, results in networks with a disparate feature space, which can protect them against UAP transfer attacks from different compressed and non-compressed models. In practice, this can be applied to design strategies to improve robustness against black-box and transfer attacks for models deployed in edge devices.
    \item We observe noticeably different outcomes between CIFAR-10 and SVHN datasets. While the same compression technique helps against targeted UAPs on the former dataset, it degrades the models' robustness on the latter. This shows that robustness to UAPs when using compression methods is really application dependant. 
    %Thus, aiming to extract general conclusions from a set of empirical evaluations in very specific datasets can lead to erroneous conclusions, yet the phenomena observed can help to understand robustness of the models across similar applications.  
    \item We perform UAP targeted attacks on both CIFAR-10 and SVHN. Our results are consistent with those for untargeted UAP attacks and reveal that Soft Filter Pruning can enhance the robustness against targeted UAPs in CIFAR-10 dataset, especially when combined with regularization techniques such as \emph{mixup} and \emph{cutout}.
    %% TODO: Add contribution about targeted UAPs?
\end{enumerate}

The rest of the paper is organized as follows. In Section 2, we describe the background on UAPs and model compression methods, and the methodology and techniques considered in our empirical evaluation. Section 3 presents our experimental results, including white-box, transfer and targeted UAP attacks. In Section 4, we discuss the main implication of our experimental results. Section 5 places this in the context of the related work. Finally, we conclude the paper in Section 6.

\section{Background \& Methodology}
\label{sec:methodology}
We describe in this section the background and methodology used in our experiments. For this, we briefly introduce adversarial examples and universal adversarial perturbations, and the compression methods we have used.

\subsection{Universal Adversarial Perturbations}
In classification problems, a neural network $F$ takes an input $x \in \mathbb{R}^n$ and outputs a class label defined by $f(x) = \argmax(F(x))$ where $F(x)$ is a probability vector $F(x) \in \mathbb{R}^m$. The classification model attains a good generalization when the model is able to correctly categorize inputs that are different from those seen during training. \citet{szegedy2014intriguing} showed that machine learning models are brittle in the presence of attackers and have proven to be particularly vulnerable to adversarial examples: malicious inputs barely indistinguishable from their legitimate equivalent that produce misclassifications in the target model. 

Let $C(x)$ be the true class label of an input $x$. An adversarial example $x' = x + \delta$ is an input in the proximity of a clean input $x$ which satisfies $f(x') \neq C(x)$. The difference $\delta = x' - x$ is referred to as an adversarial perturbation and its norm is often constrained to $\Vert \delta \Vert_p < \varepsilon$, for some $p$-norm and small $\varepsilon > 0$.

Universal Adversarial Perturbations (UAPs) can be formulated as the perturbation $\delta \in \mathbb{R}^n$ that satisfies $f(x + \delta) \neq C(x)$ for multiple inputs $x \in X$ of a benign dataset $X$. Compared to input-specific adversarial examples, UAPs exploit systemic vulnerabilities of the target model, rather than input-specific sensitivities. UAPs are also transferable across models \cite{moosavi2017universal, co2019procedural}.

In our experiments, we use the Stochastic Gradient Descent (SGD) UAP attack proposed by \citet{shafahi2019universal}. This method for generating UAPs uses Projected Gradient Descent (PGD) \cite{madry2019deep} and optimizes the adversarial perturbation over batches rather than single inputs. This is an improvement over the original iterative-DeepFool \cite{moosavi2017universal}, as SGD generates UAPs in a more efficient way, achieves higher universal evasion rates, and has convergence guarantees \cite{shafahi2019universal}. The algorithm maximizes the loss objective $\sum_i \mathcal{L}(x_i + \delta)$, where $\mathcal{L}$ is the model's training loss, $X_{\text{batch}} = \{x_i\}$ are batches of inputs, and $\delta \in \mathcal{P}$ are the valid perturbations. Updates to $\delta$ are done in mini-batches in the direction of $-\sum_i \nabla \mathcal{L}(x_i + \delta)$. For targeted UAPs, where the goal of the perturbation is to have as many adversarial examples $x + \delta$ classified as some target class $y_{\text{tgt}}$, we replace the true class labels with the target $y_{\text{tgt}}$ and then minimize the loss $\mathcal{L}$.

The perturbation constraints often take the form of an $\ell_p$-norm and appear as $\Vert \delta \Vert_{p} \leq \varepsilon$ for a chosen norm $p$ and distortion value $\varepsilon$. Note that the pixel values of modified inputs $x + \delta$ are clipped to the minimum and maximum integer values of 0 and 255 respectively.

\subsection{Compression Techniques}
%% Moved from Related Work to Background (reason: paragraph is better fit for background)
The increase in the computational power of edge devices and the advances in machine learning have led to new studies on compression techniques to reduce the memory required to deploy machine learning models in these devices whilst preserving their performance. Pruning aims to reduce the size of the DNN by removing neurons that are irrelevant or have a reduced contribution at inference time. Typically, in most practical scenarios, pruning is applied after the network is trained, and then, the compressed model is fine-tuned \cite{han2015learning, han2016deep, Zhou2016LessIM, dynamic, frankle2018lottery, liu2019rethinking}. On the other side, neural network quantization aims to reduce the memory of the deployed models by limiting the precision of the parameters of the model \cite{han2016deep, rastegari2016xnor, zhou2016dorefa, zhu2017trained}. This can be done during or after training. In this work, we evaluate three different compression techniques.

\paragraph{Post-training pruning (PP)}
This method removes the least used neurons once the DNN is trained, evaluated on a separate validation set. Thus, PP proceeds by measuring the average activation for each neuron across the different hidden-layers in the trained DNN. Then, the fraction of $p_i$ less activated neurons for each layer $i$ are removed, being $p_i$ a hyperparameter of the algorithm. This method removes the redundant activations of the network, keeping only those that add value to the classifier. The method also allows to focus and prune only specific layers, which has proven to be effective to remove some backdoors targeting DNNs by pruning only the last layer \cite{finepruning}. 

\paragraph{Soft Filter Pruning (SFP)}
This technique allows to simultaneously prune unimportant filters or neurons during training and accelerate DNN inference at test time \cite{SFP}, making it an appealing strategy not only to reduce the memory of the compressed model but also to enhance the power consumption of the deployed DNN models on the edge devices at run-time. 

This technique updates the pruned filters in the training stage allowing almost \textit{all layers} to be pruned at the same time instead of the greedy layer-by-layer pruning procedure. 

First, SFP computes the $l_p$-norm of all filters to evaluate their relative importance on the final prediction of DNNs. Filters with small $l_p$-norms will be pruned before than those of higher $l_p$-norm. In particular, SPF uses a pruning rate hyperparameter, $P_i$, to prune the lowest $N_{i+1}P_i$ filters for the $i$-th layer. Then, the selected filters are set to $0$, i.e., pruned. After the pruning step, the network is trained for one epoch, where the pruned filters are updated to non-zero due to backpropagation. SFP iterates over these three steps until convergence. At convergence, we can obtain a sparse model by removing the zeroed filters from the network.

In our experiments, in combination with this technique, we add \textit{mixup} and \textit{cutout} regularization to reduce the number of Floating-Point Operations (FLOPs), while improving the predictive accuracy. Mixup is a data augmentation method that adds convex combinations of pairs of examples and their labels to the training data \cite{zhang2018mixup}. This method constructs virtual training samples of the form $\tilde{x}=\lambda x_i + (1-\lambda)x_j$ and $\tilde{y}=\lambda y_i + (1-\lambda)y_j$ where $x_i, x_j$ and $y_i, y_j$ are raw input vectors and one-hot label encoding, respectively, drawn at random from the training set, and $\lambda\in[0,1]$. This approach \emph{teaches} the DNN to map a linear combination of inputs to a linear combination of outputs, which improves the generalization error of cutting-edge network architectures, alleviates the memorization of corrupt labels, and helps combat the sensitivity to adversarial examples and the instability in adversarial training \cite{zhang2018mixup}. The cutout technique is also a data augmentation scheme which adds occluded versions of existing samples to the training set \cite{devries2017improved}. Cutout applies a fixed-sized zero-mask to a random square region of each input image during training, preventing the DNN from relying on specific visual features. This approach improves robustness and yields higher test accuracy levels \cite{devries2017improved}.

\paragraph{Additive Powers-of-Two Quantization (APoT)}

This technique proposed by \citet{quant_apot} aims to reduce the memory of DNNs by quantizing their weights and activations. Quantization has been commonly used as an alternative to network pruning for deploying machine learning models in edge devices. APoT is a non-uniform quantization scheme applied at training time in which each quantization level of APoT is the sum of a set of powers-of-two terms. This approach has a Reparameterized Clipping Function (RCF) that computes a better-defined gradient for the optimization of the clipping threshold. In addition, the algorithm normalizes the weights in the forward path, which results in a more stable and consistent training. We consider this quantization method because it reaches state-of-the-art accuracy on benchmark computer vision datasets and decreases 22\% of the computational cost when compared to uniform quantization \cite{quant_apot}.

\section{Experiments}
\label{sec:setup}
In this section we present our experimental results. We first describe our experimental setup, and then, we present our results, measuring the robustness of compressed models to UAPs, under white-box settings, and transferability of UAPs to compressed models.

\subsection{Experimental Setup}

\paragraph{Models \& Datasets}
We perform our experiments comparing the \emph{full precision (Full)} and structured compression methods on DNN models trained on CIFAR-10 \cite{krizhevsky2009learning} and SVHN \cite{netzer2011reading} datasets. The images in both datasets have the same dimensionality ($32 \times 32 \times 3$ pixels for each image). We use the ResNet-18 \cite{he2016deep} model architecture for the full precision model on both datasets. All models are trained with SGD, an initial learning rate of 0.1, and a step-wise decay by a factor of 10 after 80 and 120 epochs on CIFAR-10, and 10 and 15 on SVHN dataset. The momentum coefficient is set to 0.9 with a learning rate of 0.0001 and a batch size of 250.%The codebase is implemented in PyTorch and will be made available at \url{https://github.com/mataach}.

\begin{table}[t]
\centering
\begin{tabular}{lrr}
\toprule
        Model &  CIFAR-10 &  SVHN \\
\midrule
      Full &      94.02 &      \textbf{97.26} \\
       SFP &      79.51 &      96.81 \\
     SFP+M &      86.09 &      96.82 \\
     SFP+C &      83.54 &      97.11 \\
       PP2 &      91.65 &      93.67 \\
       PP3 &      93.52 &      93.61 \\
       PP4 &      \textbf{94.09} &      94.14 \\
        Q2 &      90.47 &      95.46 \\
        Q3 &      92.57 &      95.31 \\
        Q4 &      92.20 &      94.99 \\
\bottomrule
\end{tabular}
\caption{Clean model accuracy (in \%) for ResNet18. Models with highest accuracy for each dataset are highlighted.}
\label{table:clean}
\end{table}

\paragraph{Compression Techniques} 
For Soft Filter Pruning (SFP) we set the pruning rate, $P_i$, to 0.9, i.e., 10\% of the filters are pruned from each layer during training, as in \cite{SFP}. For the two regularization methods, mixup and cutout, we use the default values proposed by \citet{zhang2018mixup} and \citet{devries2017improved}. For mixup interpolation we use an $\alpha$ coefficient of $1.0$ and for cutout regularization we use a single hole per image with a default length of 16. For Post-training Pruning (PP), we fine-tune the ``Full" model after it is trained. We set the compression parameter, $p_i$, to 0.7 for each hidden layer block $i = 2, 3, 4$, i.e., 30\% of the activations of the convolutional blocks $(128\times16\times16)$, $(256\times8\times8)$, and $(512\times4\times4)$ are pruned. We perform our evaluation pruning only one block at a time, denoting the corresponding models as PP2, PP3, and PP4 for blocks 2, 3, and 4, respectively. We choose a coefficient $p_i$ which does not harm the clean Full model accuracy by more than 5\%. For APoT quantization we quantize the weights and activations of the hidden layers to 2, 3, and 4 bits, always keeping the first and last layer of the network with 8 bits.

The following is a complete list of the models considered according to the compression method:
\begin{itemize}
	\item \textbf{Full precision:} Full (using a standard ResNet-18).
	\item \textbf{Soft Filter Pruning:} SFP, SFP+M (mixup), SFP+C (cutout).
	\item \textbf{Post-training Pruning:} PP2, PP3, PP4 (pruning blocks 2, 3, and 4 respectively).
	\item \textbf{APoT quantization:} Q2, Q3, Q4 (using 2, 3, and 4 bits respectively).
\end{itemize}

The clean accuracy for each of the trained models is shown in Table~\ref{table:clean}. We observe that, in general, models are able to achieve better accuracy on SVHN. For CIFAR-10, we observe that SFP significantly drops the performance of the model, even when using mixup and cutout regularization. In contrast, PP barely degrades the performance in CIFAR-10 compared to the full model. In the case of SVHN, SFP outperforms the other compression methods, although the drop in performance in all cases is not significant.

\paragraph{Attacks \& Metrics}
We use SGD \cite{shafahi2019universal} to generate the UAP attacks. In this scenario, the attacker needs access to a model's internal parameters and clean data. We consider two scenarios: (1) \textit{untargeted} UAPs that maximize the overall model's classification error, and (2) \textit{targeted} UAPs that force the model to classify input samples as the desired target class. To measure the attacker's success for these two goals, we define the \emph{Universal Evasion Rate} (UER) and the \emph{Targeted Success Rate} (TSR) of a UAP perturbation $\delta$ over the dataset, with the former being used for untargeted UAPs and the latter used for targeted UAPs. These metrics for $\delta$ over dataset $X$ are defined as follows:
\begin{equation*}
\text{UER}(\delta) = \vert \{x \in X : \argmax F(x + \delta) \neq C(x) \} \vert \cdot \frac{1}{\vert X \vert}
\end{equation*}\label{eq:uer}
\begin{equation*}
\text{TSR}(\delta, y_{\text{tgt}}) = \vert \{x \in X : \argmax F(x + \delta) = y_{\text{tgt}} \} \vert \cdot \frac{1}{\vert X \vert}
\end{equation*}\label{eq:tgt} where $\Vert \delta \Vert_{\infty} \leq \varepsilon$ is the perturbation constraint, $\vert X \vert$ is the cardinality of $X$, and $y_{\text{tgt}}$ denotes the target class label for the targeted UAP. We use the $\ell_{\infty}$-norm perturbation constraint, as it is the standard in the UAP and computer vision literature. It ensures that the perturbations are small enough and do not substantially change visual appearance of the image.

In our experiments, for untargeted UAPs under white-box settings, we explored values for $\varepsilon$ from $2$ to $12$. For transfer attacks with untargeted UAPs and for targeted white-box UAPs we set $\varepsilon = 10$

% In our experiments we use the following settings to constrain the adversarial perturbations:
% \begin{itemize}
%     \item Untargeted UAP white-box attack for $\varepsilon = 2$ to $12$
%     \item Untargeted UAP transfer attack at $\varepsilon = 10$
%     \item Targeted UAP white-box attack at $\varepsilon = 10$
% \end{itemize}

For the untargeted white-box attacks, we evaluate the effectiveness of the UAPs on the models that they were generated from (e.g. the UAP generated from SFP is then evaluated on SFP). For the transfer attack, we evaluate each of the generated untargeted UAPs across all the other models. The model that the UAP is generated from is referred to as the ``Attack Source''. For the targeted white-box attack, we generate a targeted UAP for each of the 10 classes on both datasets for a given model, and then we evaluate these UAPs on that same model.

Measuring the robustness to UAPs in the white-box setting for both untargeted and targeted UAPs gives us a good idea of how robust the model is to attacks generated from itself. Measuring the robustness to transfer attacks is useful to understand the viability of black-box attacks where the attacker is in possession of a similar model, whether in its compressed or in its full precision form. 

Such attacks are realistic for DNNs deployed on resource constrained devices. For example, an attacker can physically acquire a device and extract the compressed model. From there, the attacker can generate UAPs to attack the full precision model that could possibly be deployed online. Similarly, an attacker that has access to the full precision model online, can use it to mount attacks on the compressed model or, alternatively, build a surrogate model and try to exploit the attack's transferability. In the rest of this section, we discuss our results for each of the three main experiments.

\begin{figure}[t]
\centering
\includegraphics[width=\columnwidth]{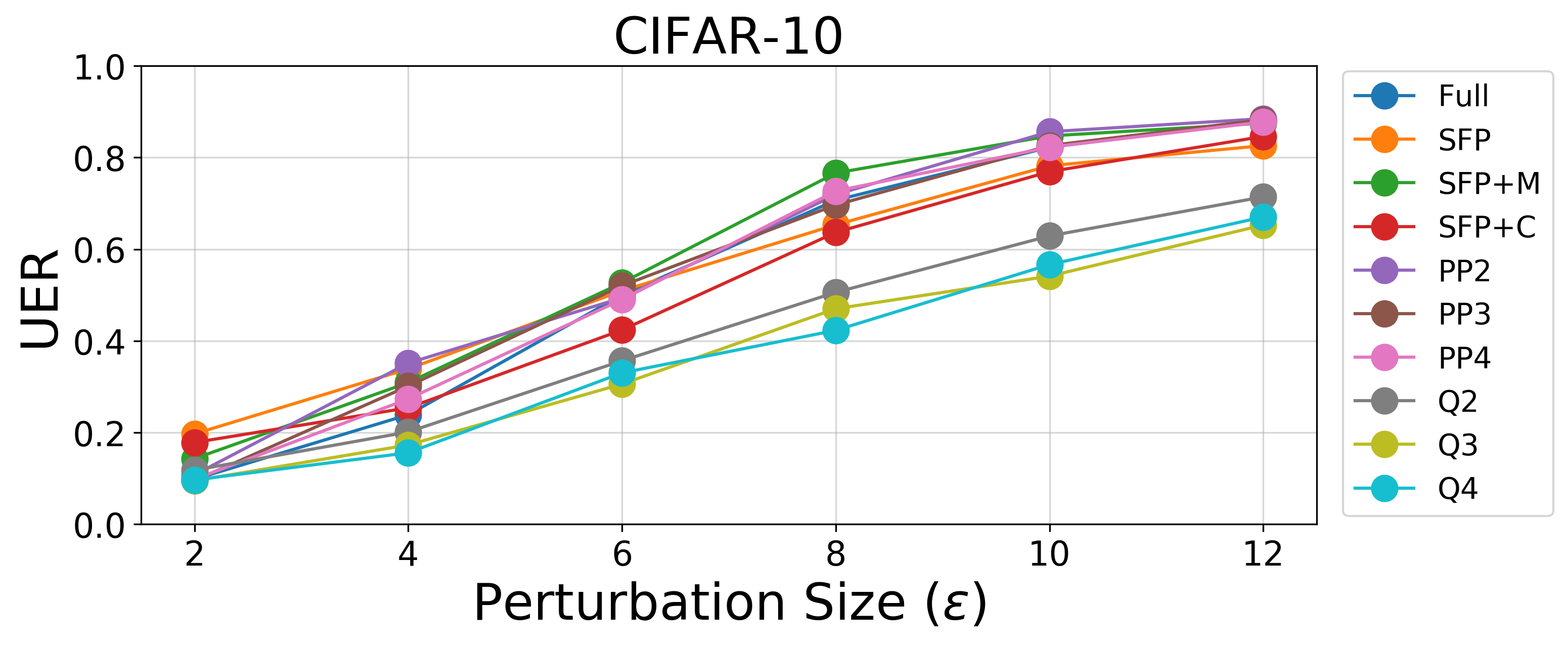}
\includegraphics[width=\columnwidth]{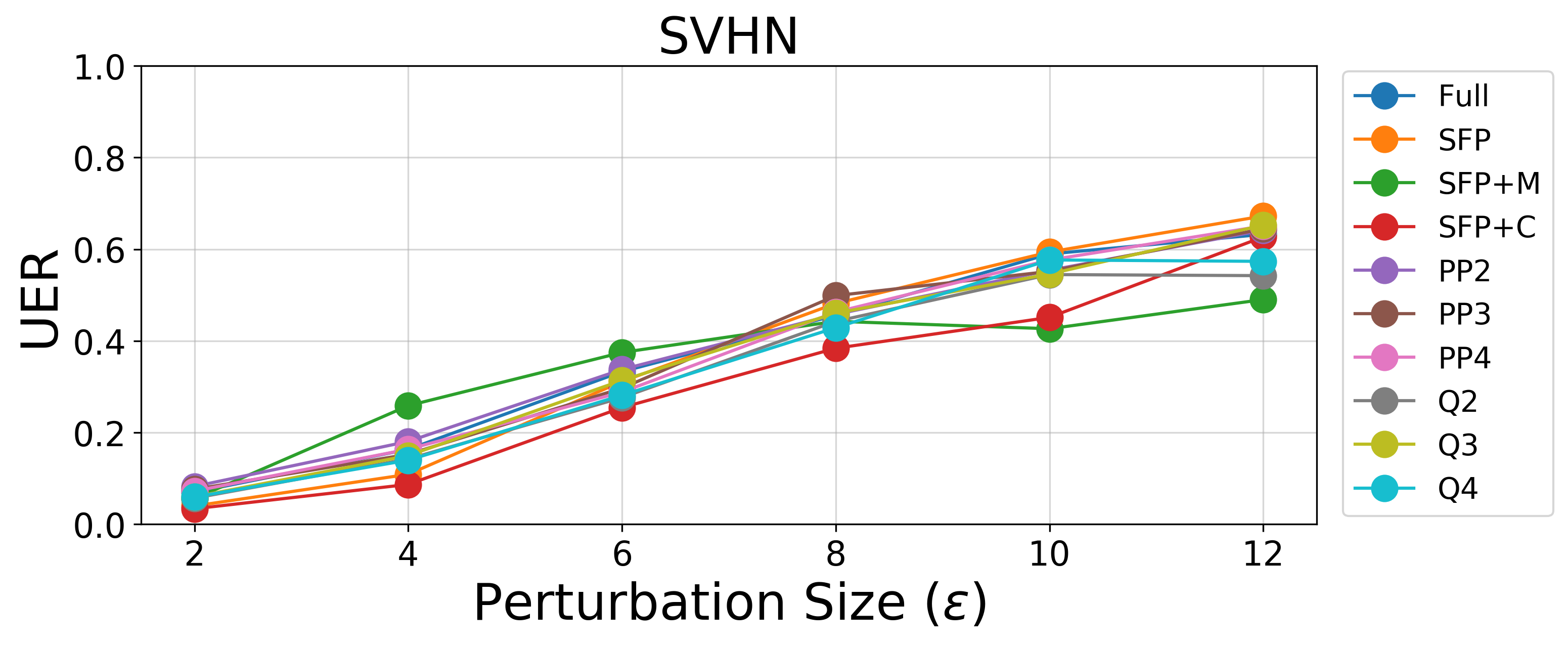}
\caption{UER for untargeted white-box UAPs at various perturbation levels $\varepsilon$.}
\label{fig:whitebox}
\end{figure}

\begin{figure*}[t]
\centering
\includegraphics[width=1.03\columnwidth]{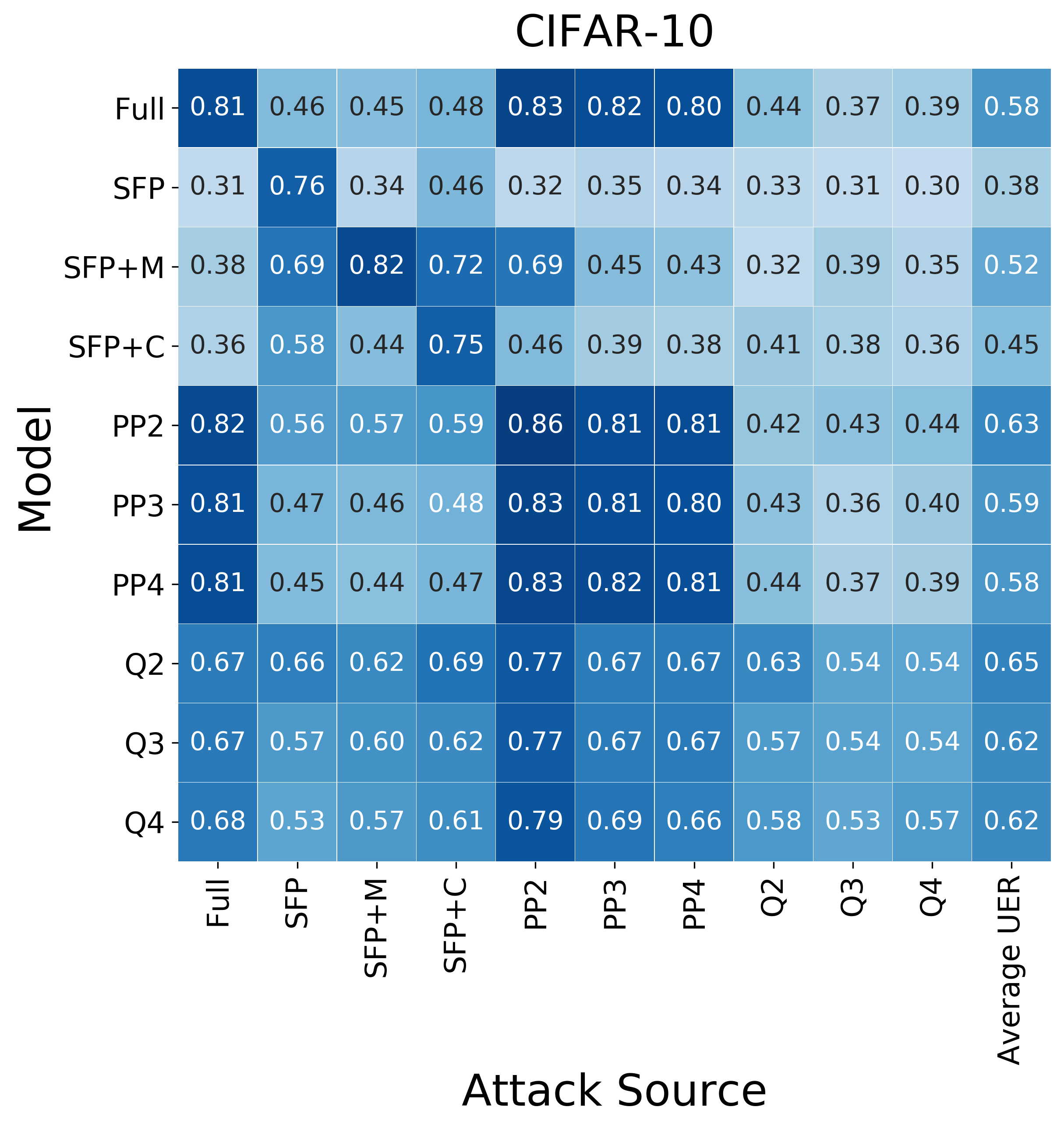}
\includegraphics[width=1.03\columnwidth]{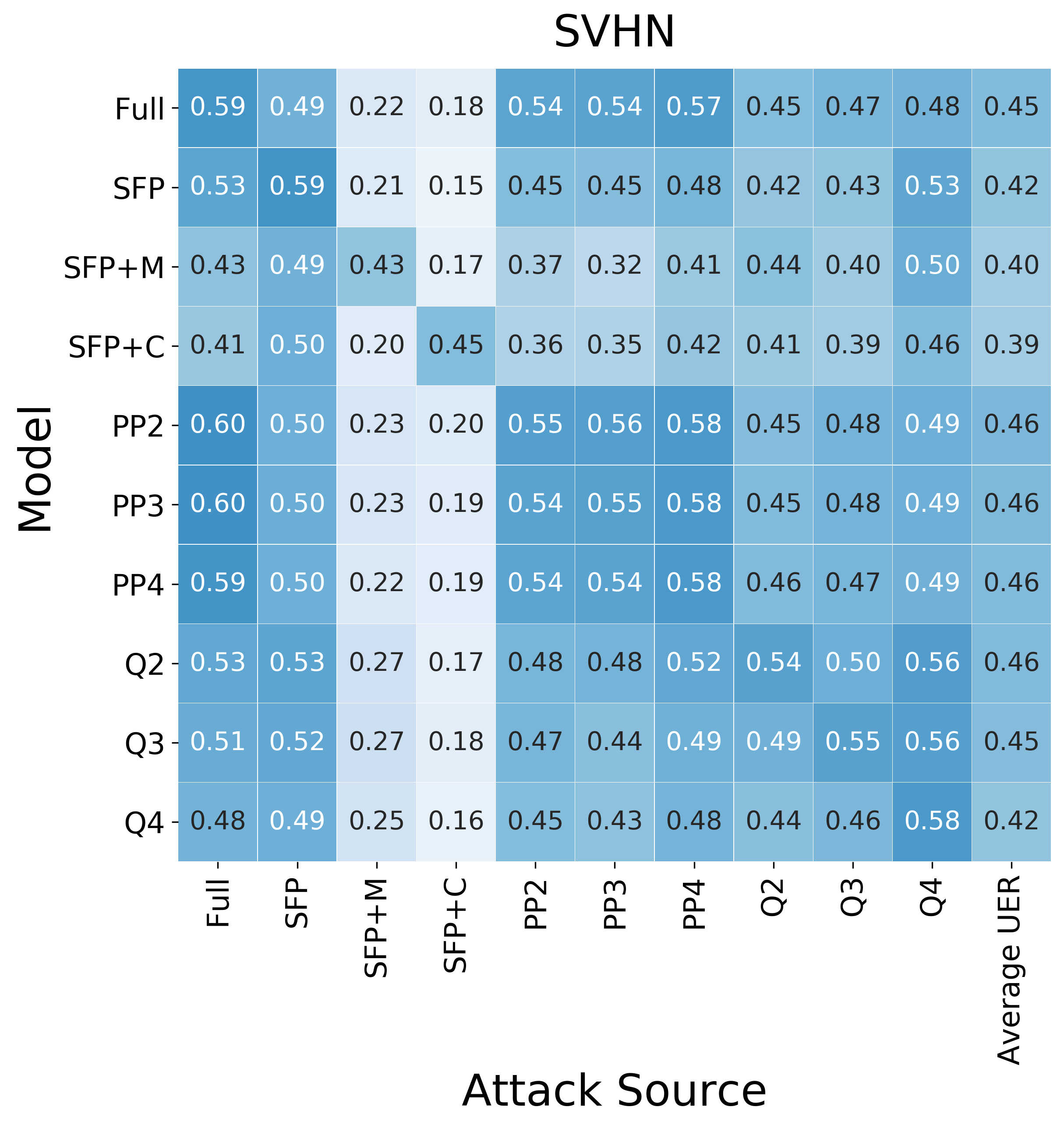}
\caption{UAP transferability on ResNet18 models with different compression methods for both datasets. Values shown are UERs of untargeted UAPs from attack source with maximum perturbation value $\varepsilon = 10$.}
\label{fig:transfer}
\end{figure*}

\subsection{White-box Robustness}
We evaluate the UER for untargeted white-box UAPs at different perturbation levels $\varepsilon$. We use the Full model as the reference, and note that, as expected, errors increase as the adversarial samples become stronger. When compared to this reference, compressed models do not appear to be significantly more robust. The only noticeable increase in robustness was for the quantized models on CIFAR-10, which display a lower average UER as $\varepsilon$ increases. The same increase in robustness is not evident for the models trained on SVHN. 

As we discuss later, it is possible that the robustness gained from quantization is a result of gradient masking: the quantized model might be smoothing the decision surface and reducing gradients used by adversarial crafting in small neighborhoods \cite{papernot2017practical, gupta2020improved}. In practice, these models are still vulnerable to adversarial examples that affect a smooth version of the same model. Therefore, to determine whether gradient masking occurs in our experiments, we check later on the transferability of UAP attacks from the full precision model to these quantized models.

Interestingly, models trained on CIFAR-10 display a much higher UER than models trained on SVHN for the same perturbation size $\varepsilon$. At the same time, SVHN models reach a higher baseline accuracy than CIFAR-10 models (see Table~\ref{table:clean}). We hypothesize that this difference comes from the different structure of the two datasets presented: CIFAR-10 contains a wide variety of objects with different shapes and colours, whereas the variety of shapes for the digits in SVHN dataset is reduced. This is reflected on the higher performance of the models for SVHN, the classification problem is ``easier" to solve. The better defined structure of the objects in SVHN also limits the effectiveness of the UAP attacks compared to CIFAR-10. However, this does not necessarily mean that the models trained for SVHN are more robust to other type of perturbations, for instance, input-specific adversarial examples. 

% We believe that more accurate DNNs learn a more separable feature space which makes the SVHN models generalize better than the CIFAR-10, leading to both higher baseline accuracy and improved robustness. 

\begin{figure*}[!ht]
\centering
\includegraphics[width=0.65\columnwidth]{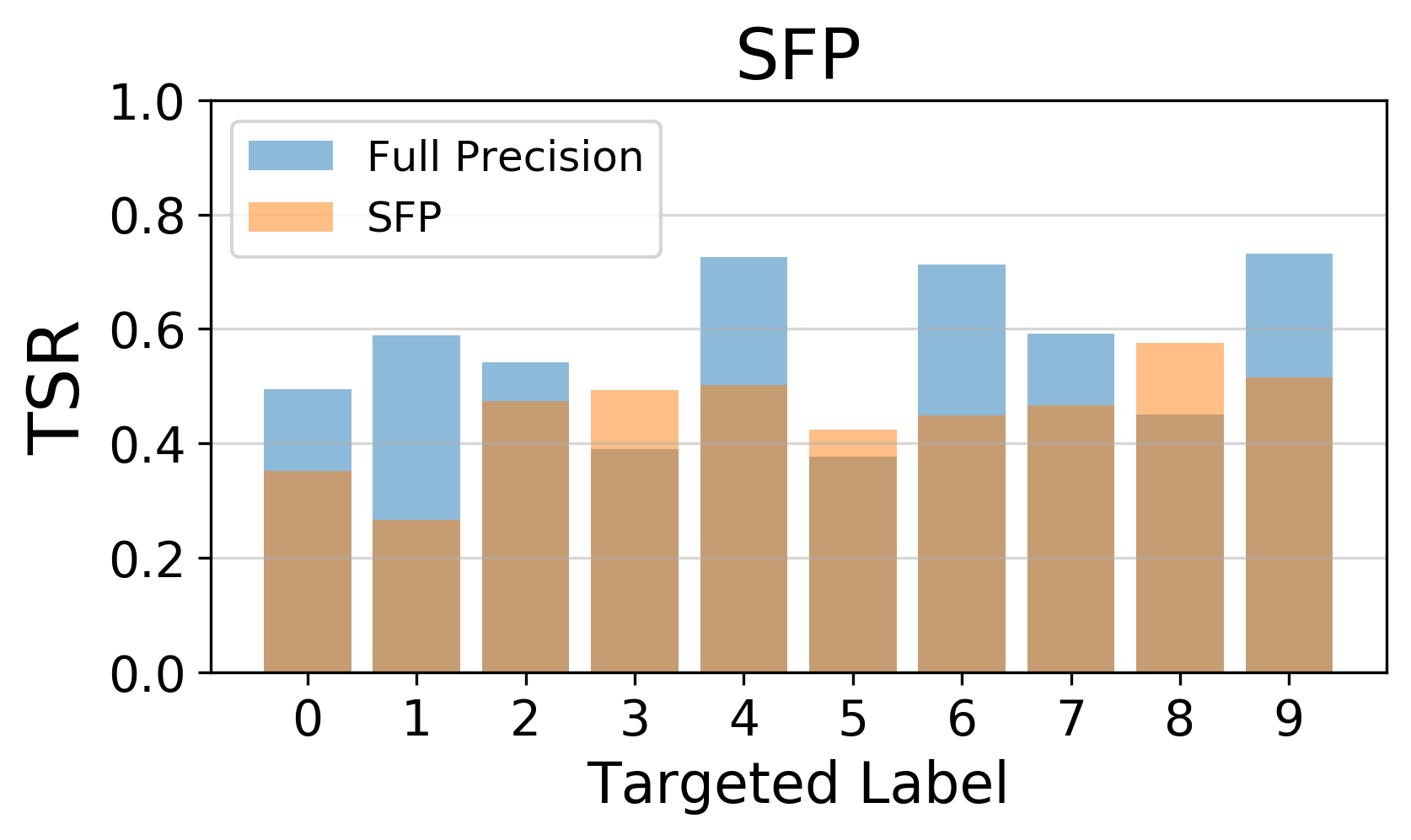}
\includegraphics[width=0.65\columnwidth]{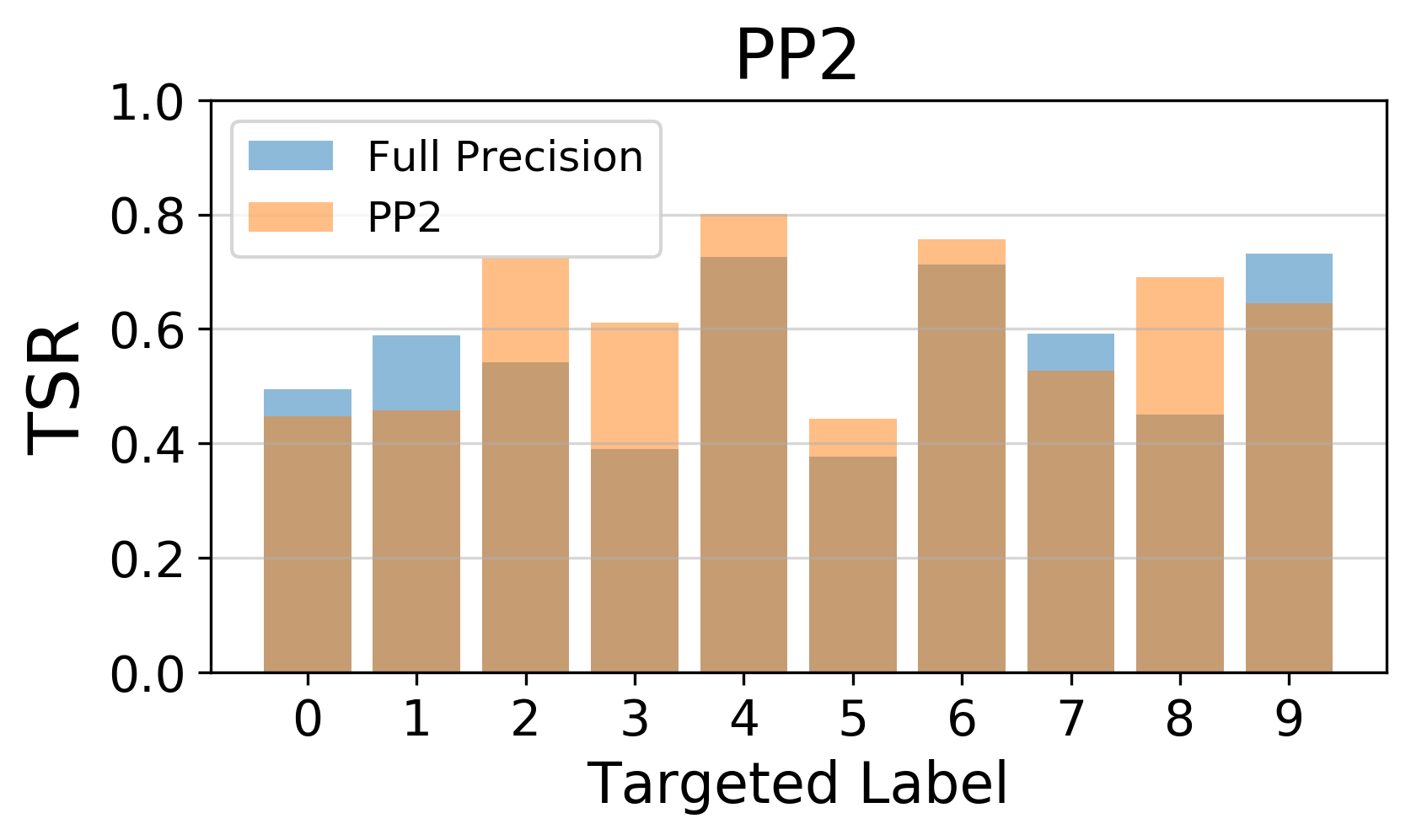}
\includegraphics[width=0.65\columnwidth]{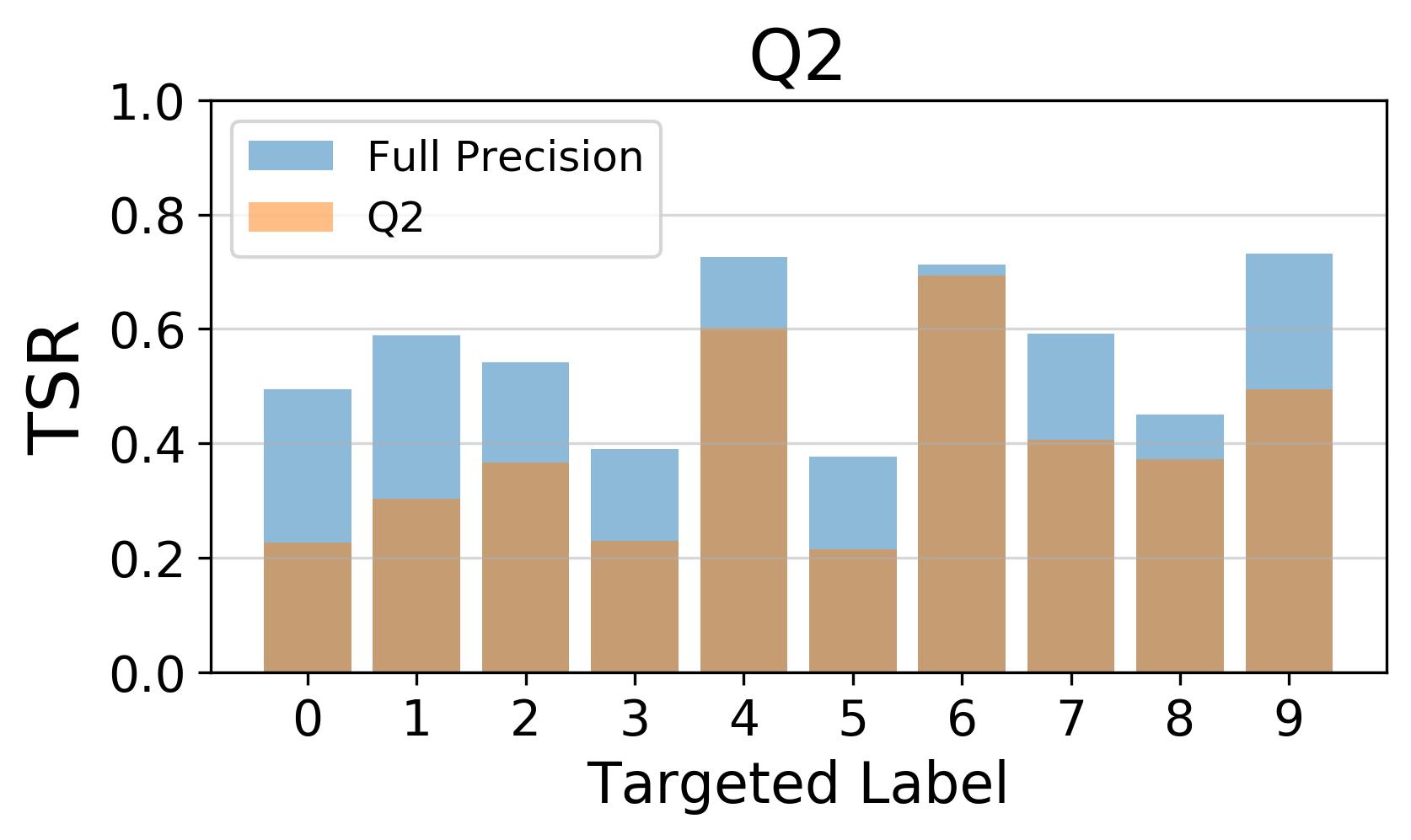}\\
\includegraphics[width=0.65\columnwidth]{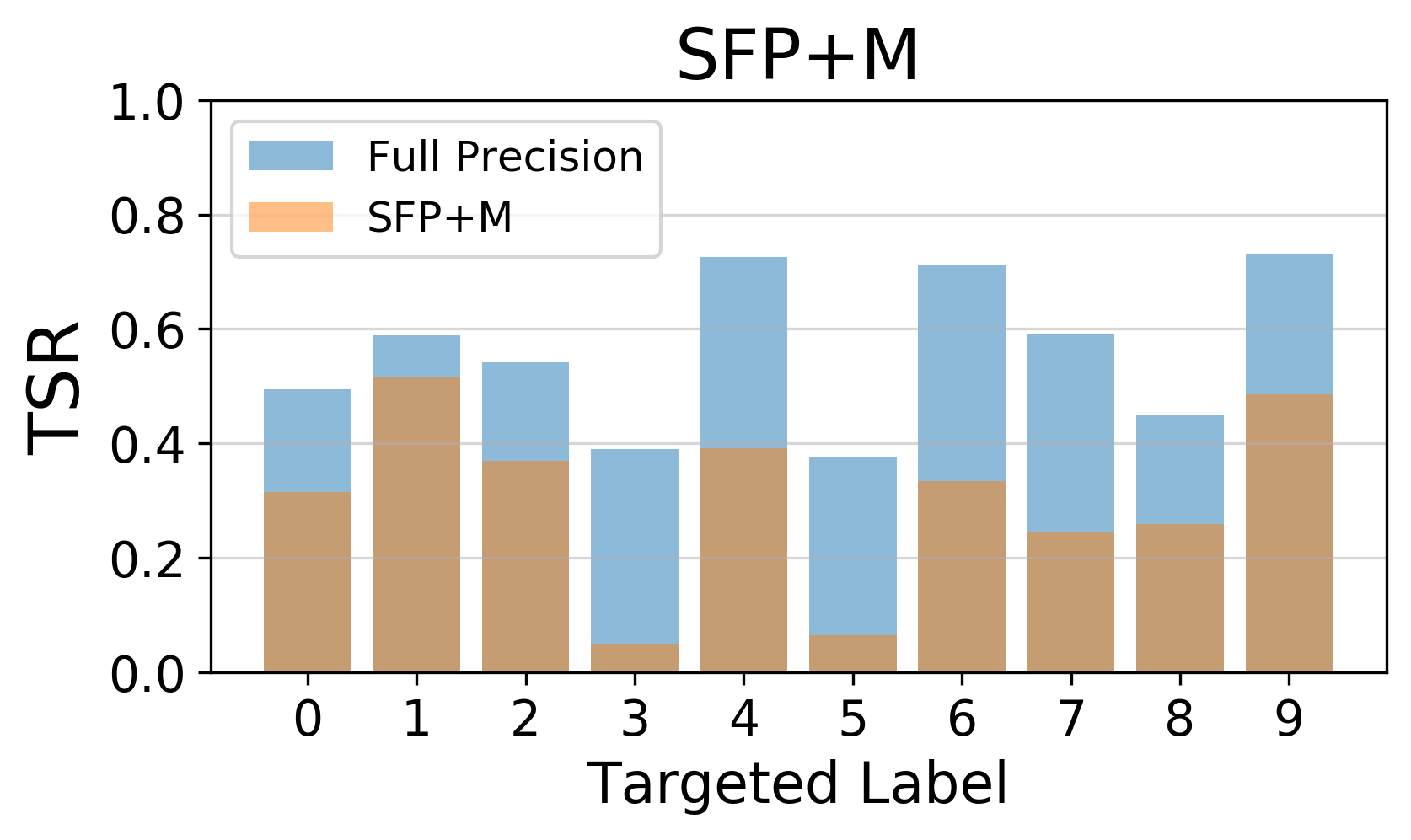}
\includegraphics[width=0.65\columnwidth]{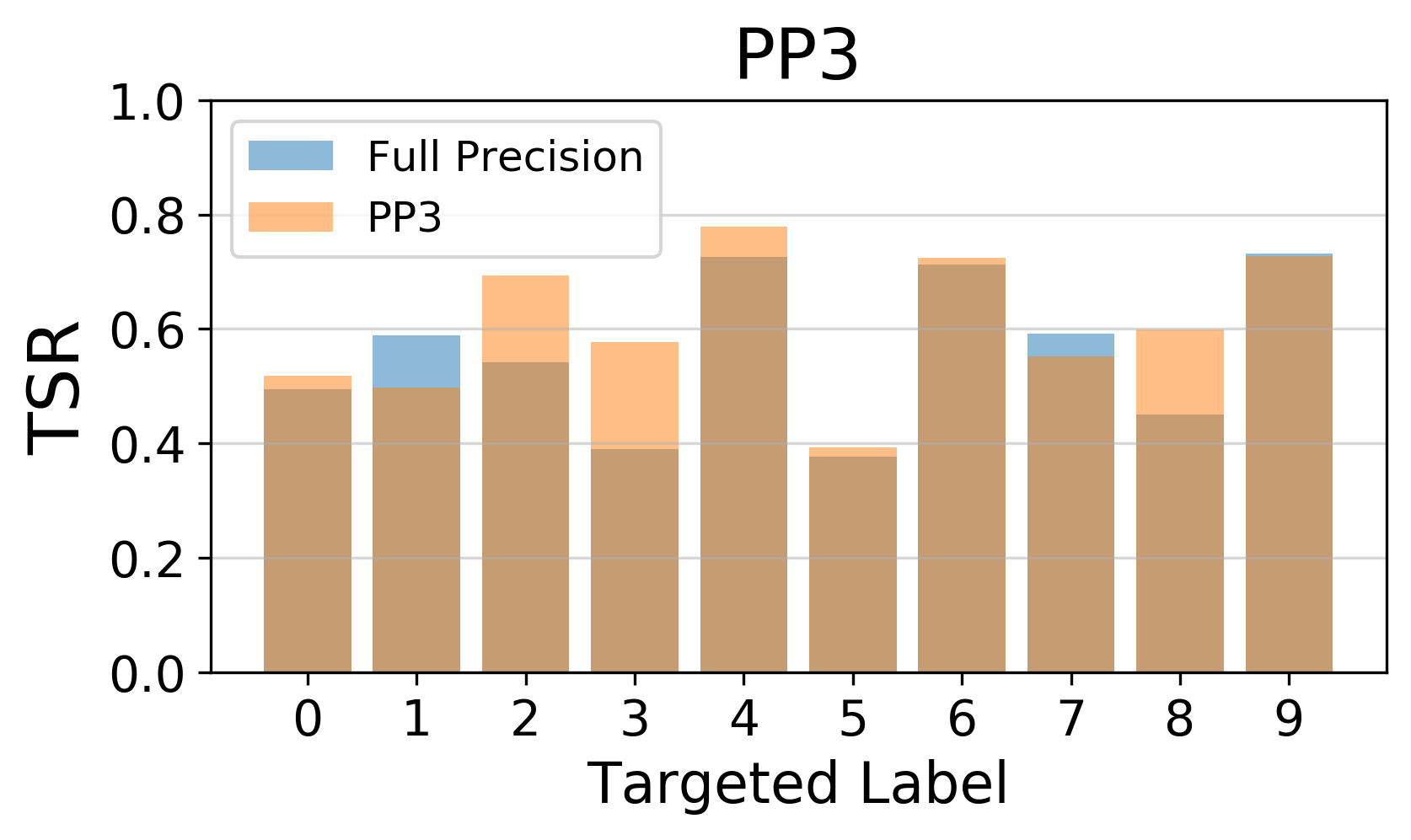}
\includegraphics[width=0.65\columnwidth]{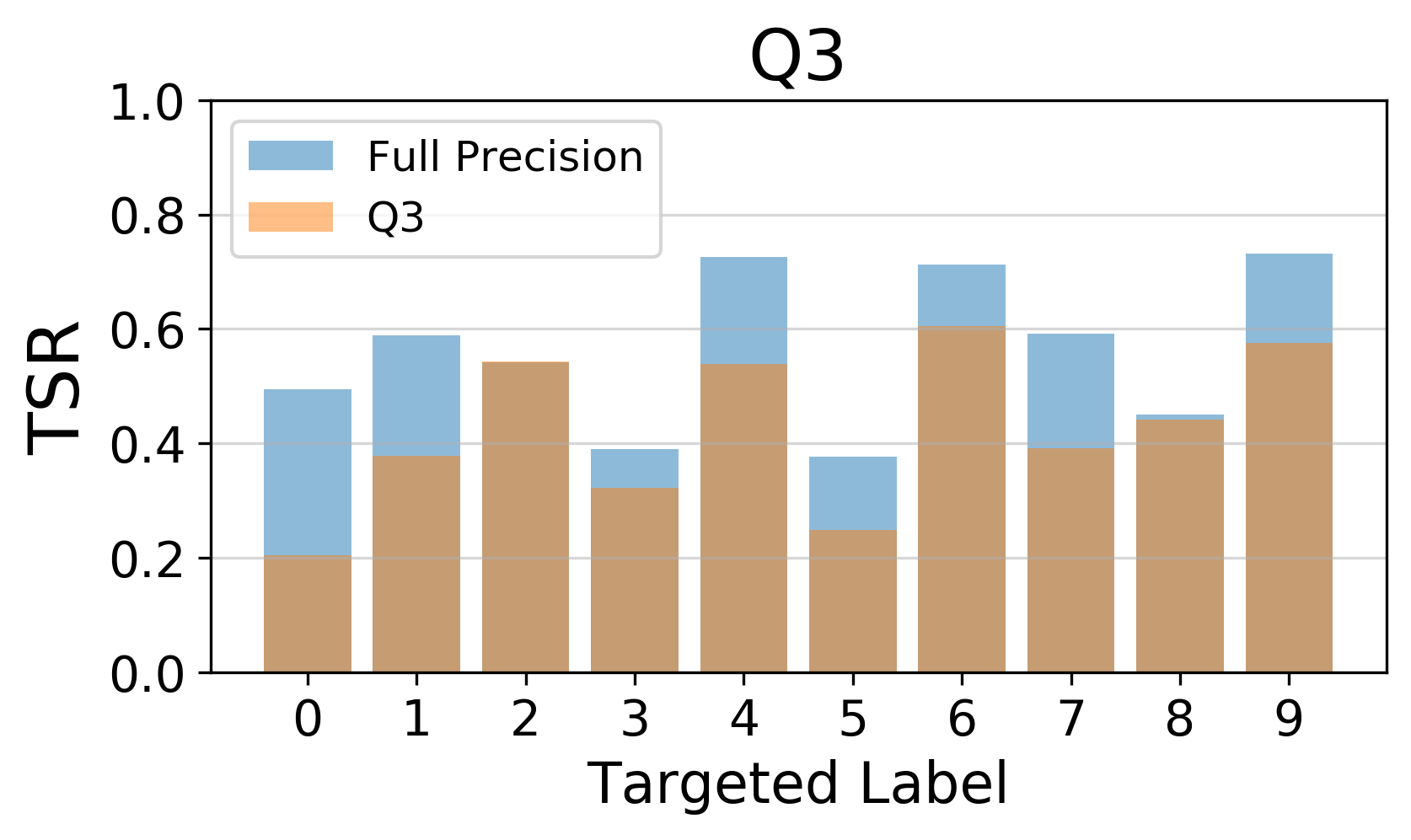}\\
\includegraphics[width=0.65\columnwidth]{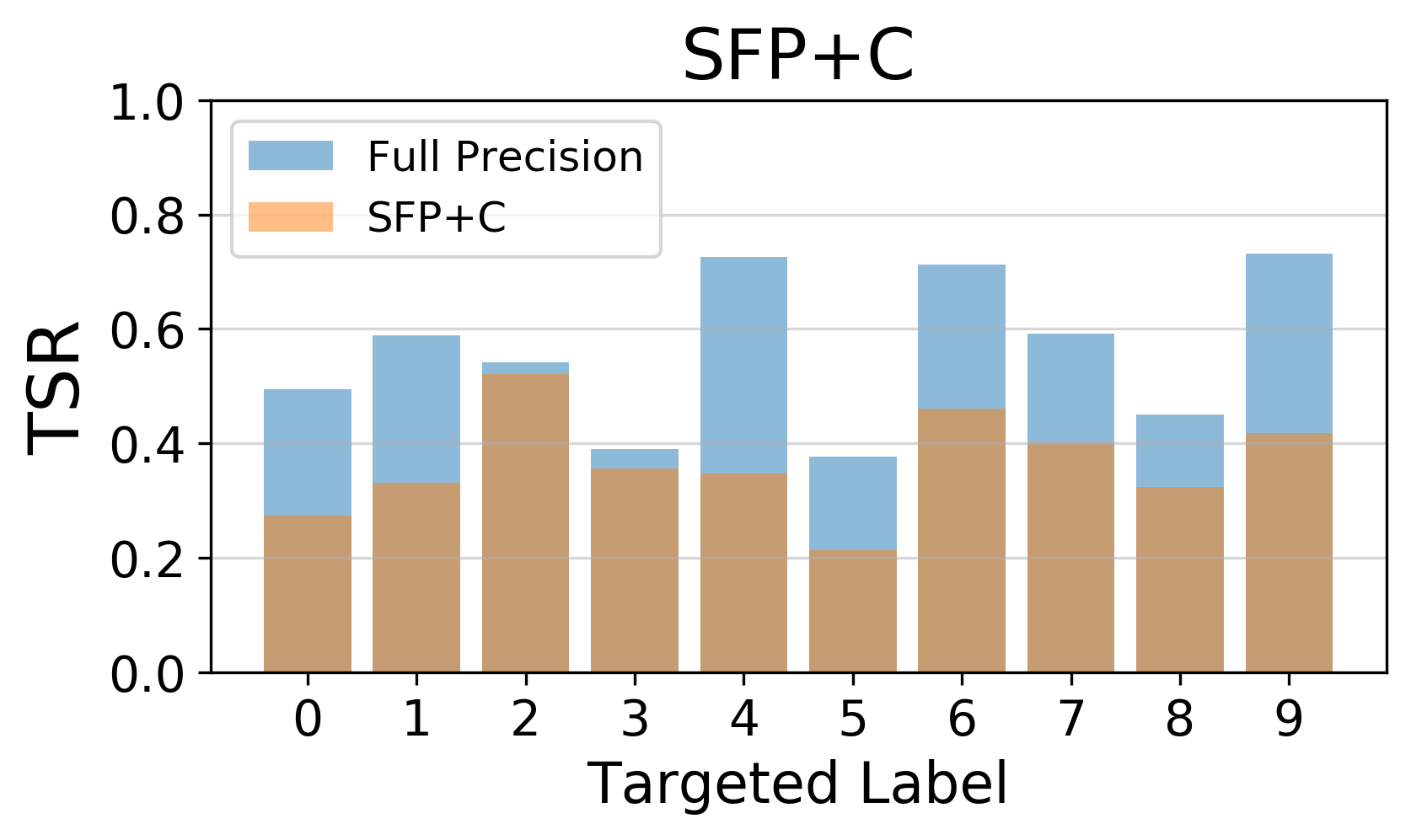}
\includegraphics[width=0.65\columnwidth]{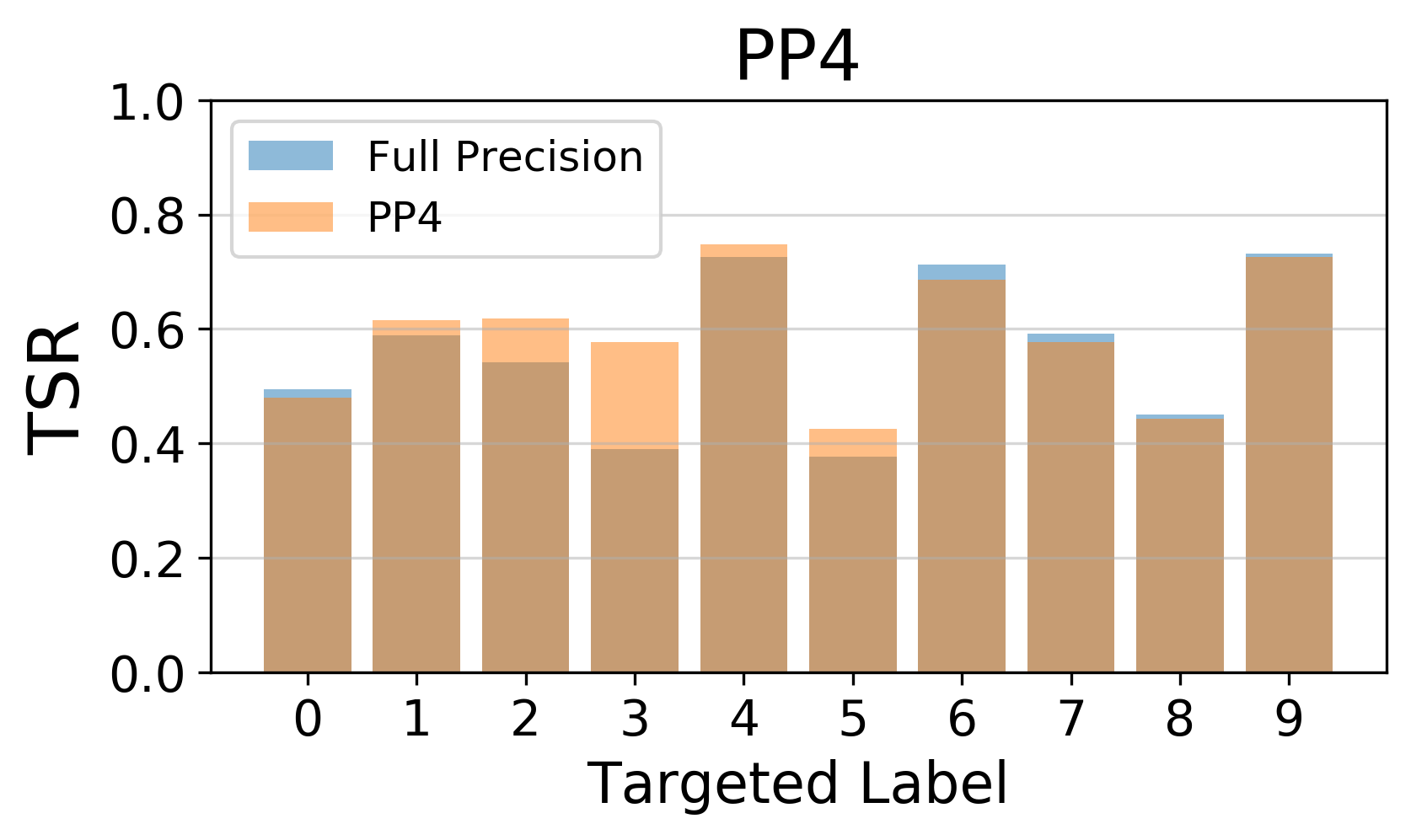}
\includegraphics[width=0.65\columnwidth]{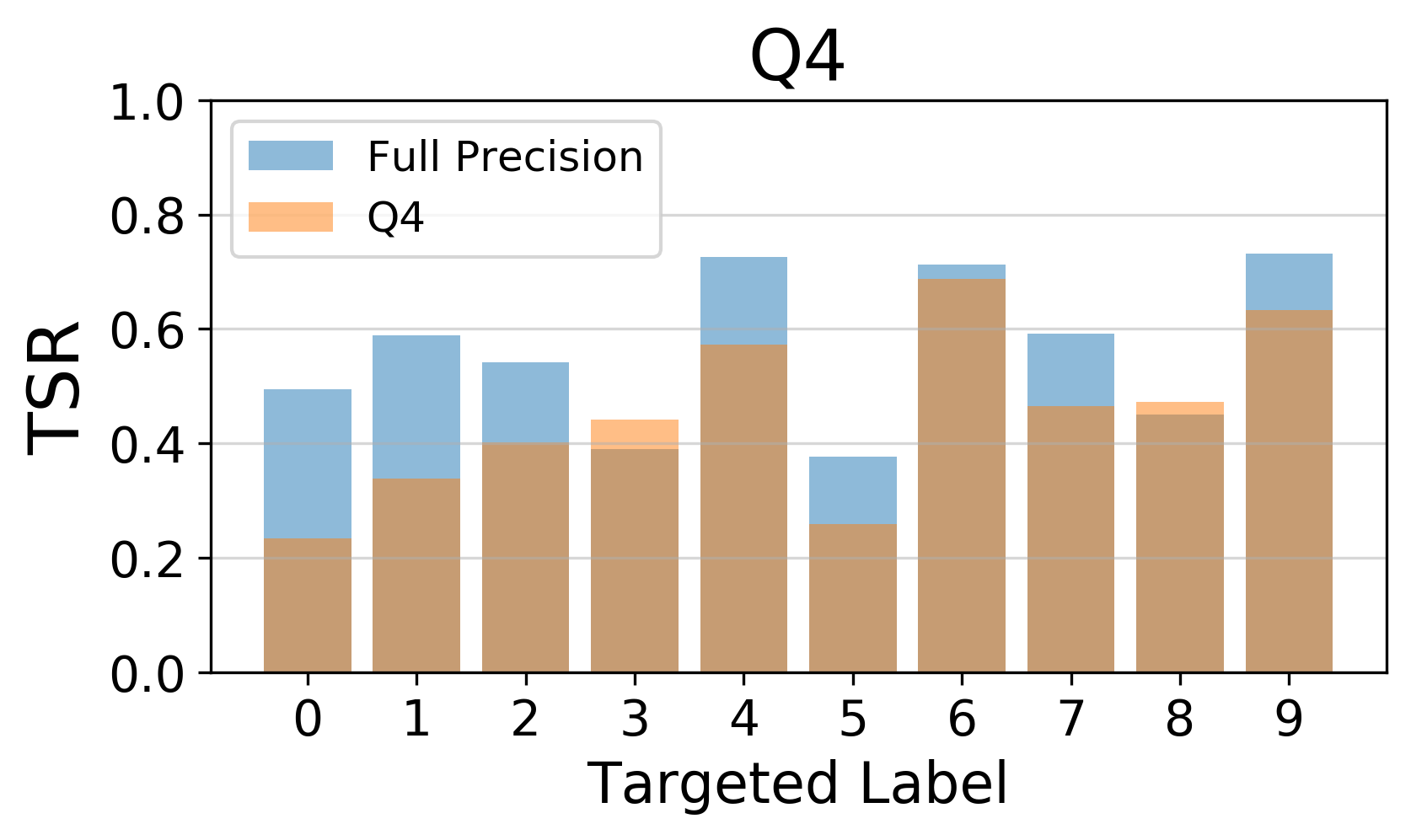}\\
\caption{Targeted UAP results for models on CIFAR-10. The TSR of targeted UAPs on the full precision model is shown (in blue) to compare with the TSR of the corresponding compressed model (in orange). Seeing more blue indicates that the compressed model improves targeted UAP robustness, while seeing more orange indicates that the compressed model is more vulnerable to targeted UAPs.}
\label{fig:targeted_1}
\end{figure*}

\begin{figure*}[!ht]
\centering
\includegraphics[width=0.65\columnwidth]{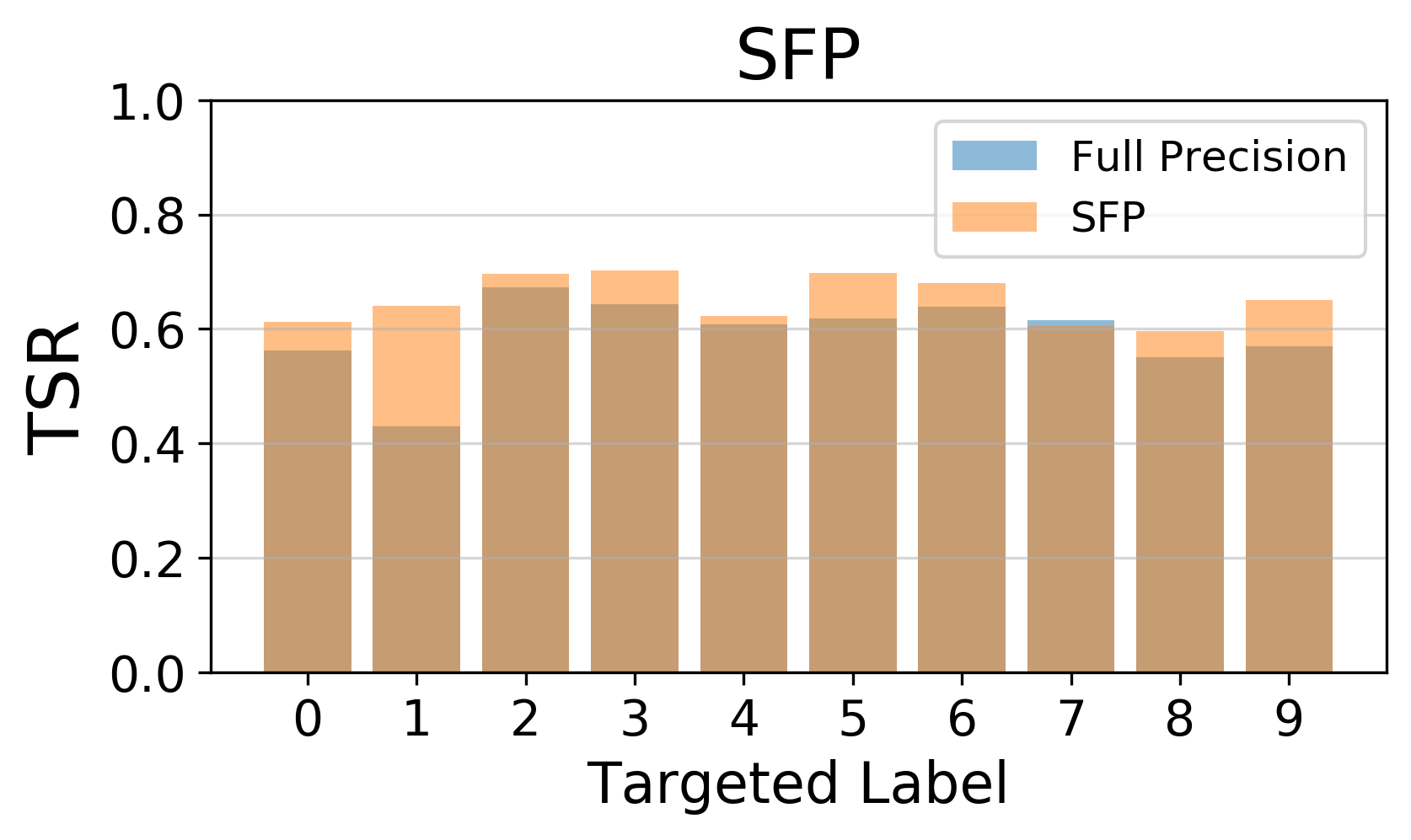}
\includegraphics[width=0.65\columnwidth]{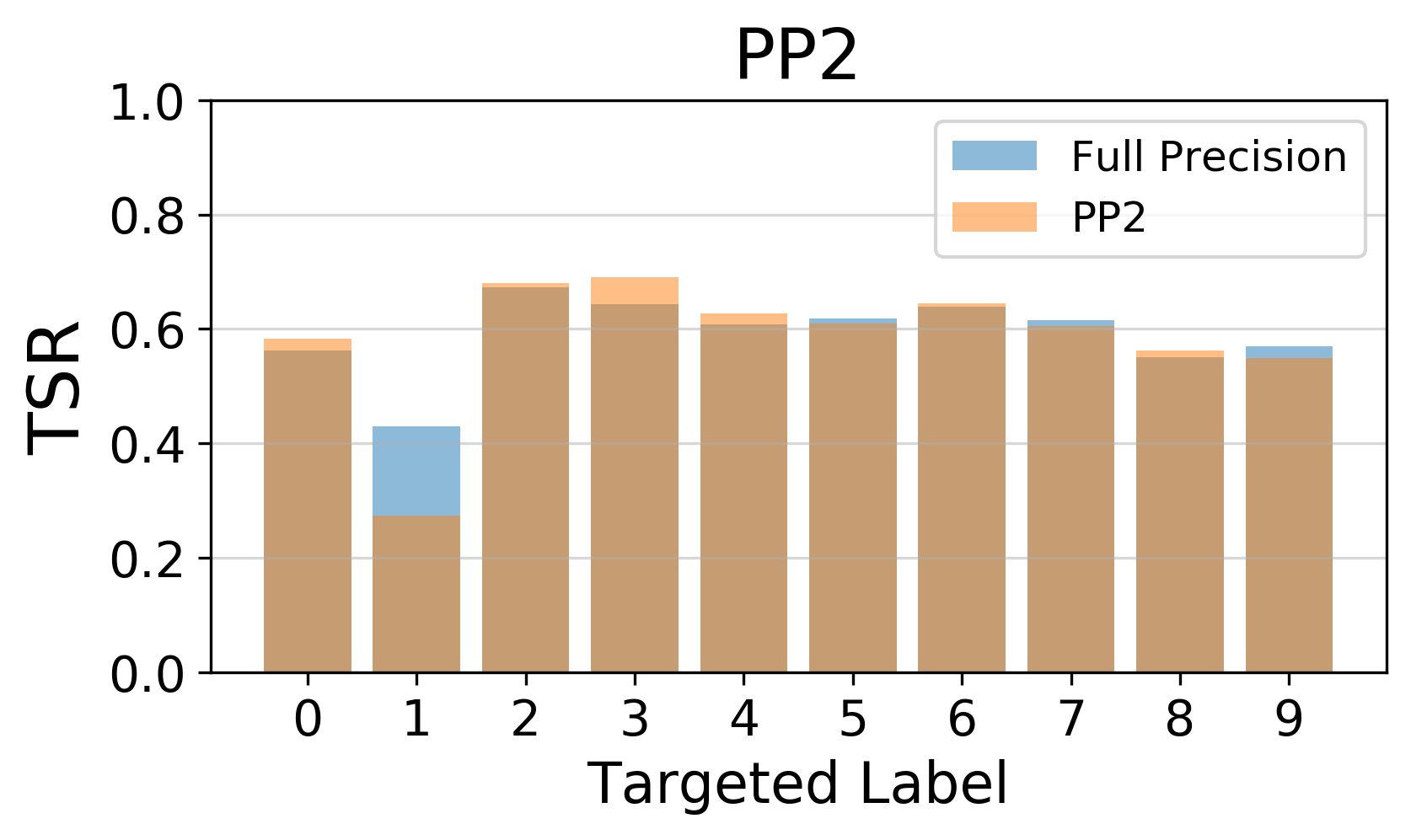}
\includegraphics[width=0.65\columnwidth]{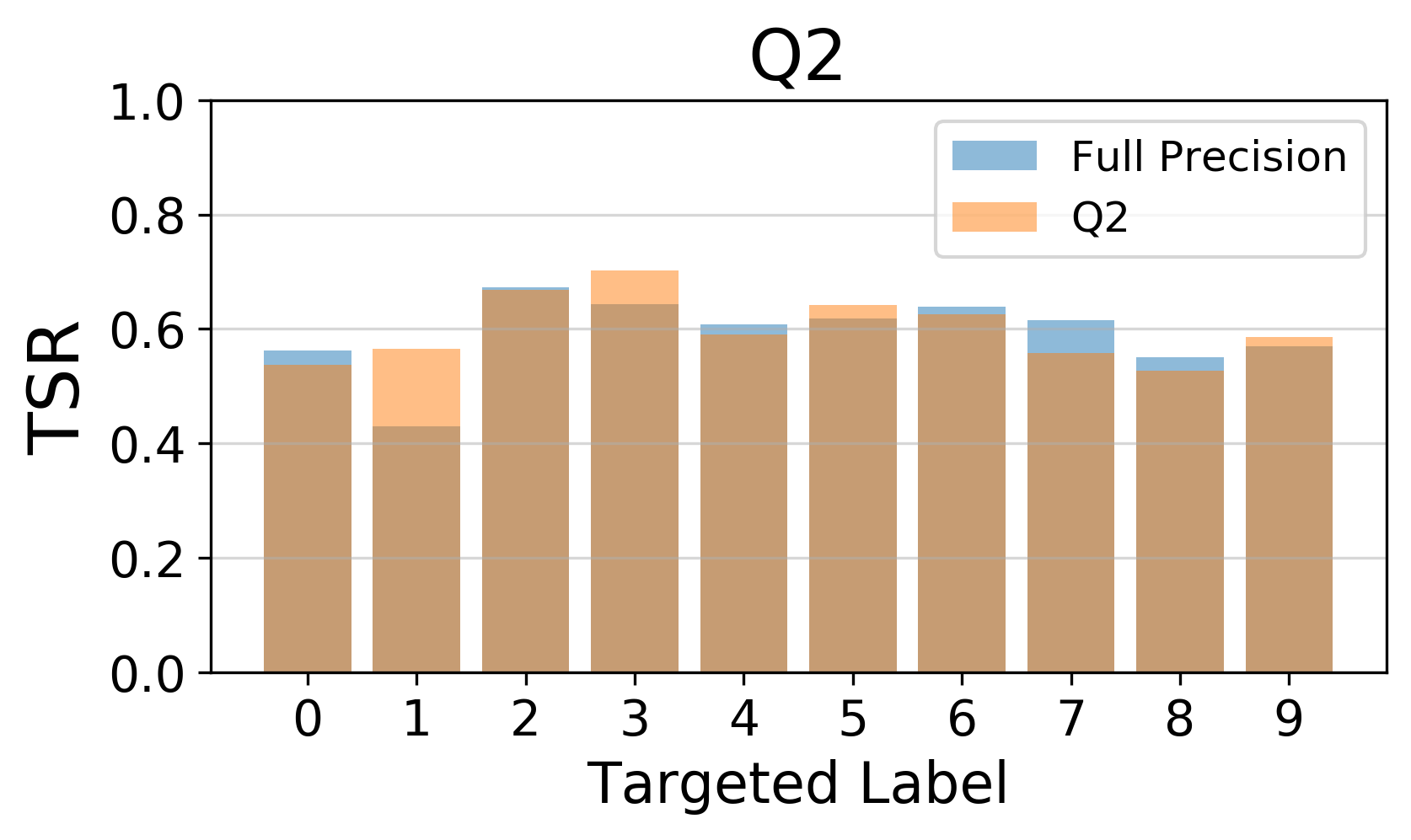}\\
\includegraphics[width=0.65\columnwidth]{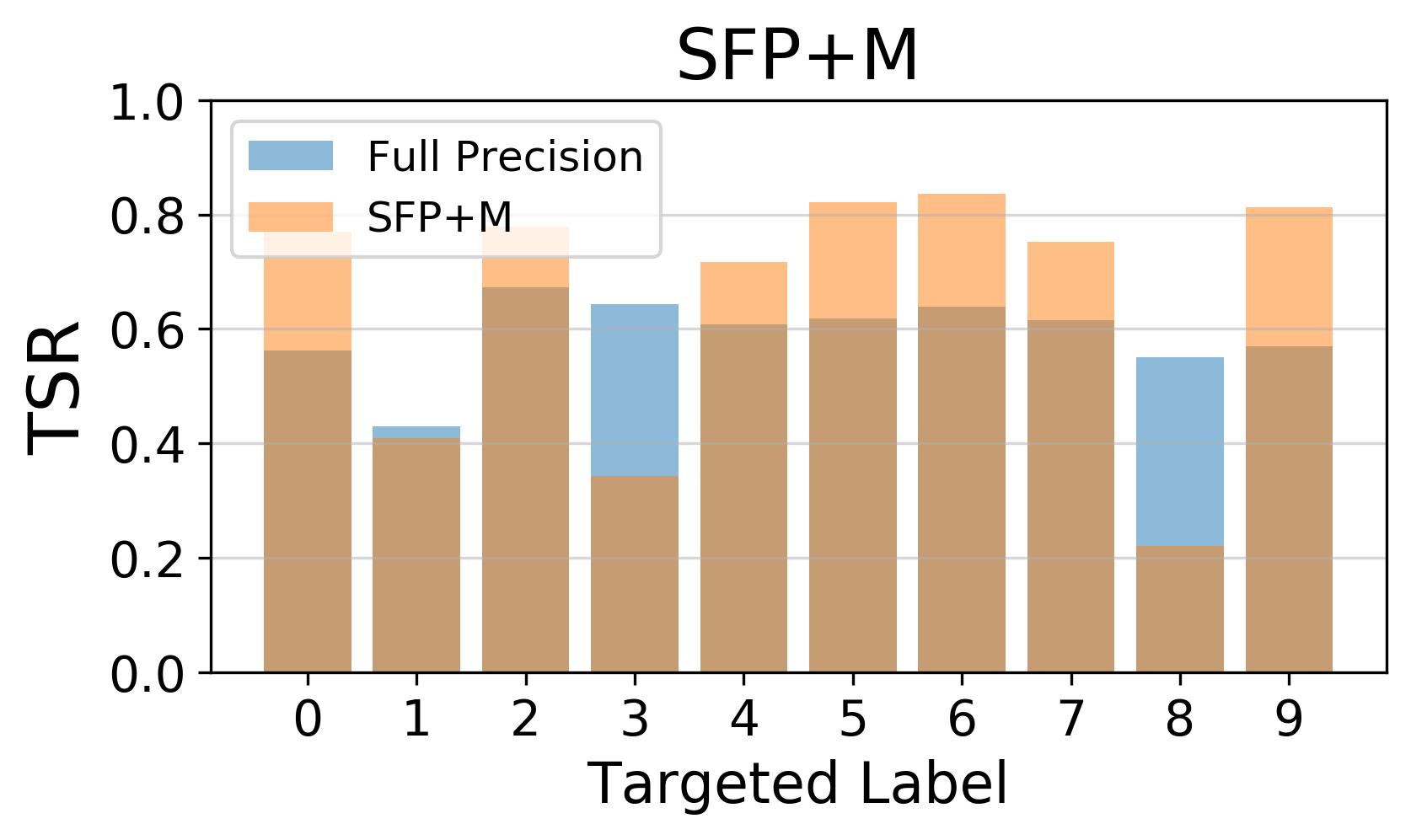}
\includegraphics[width=0.65\columnwidth]{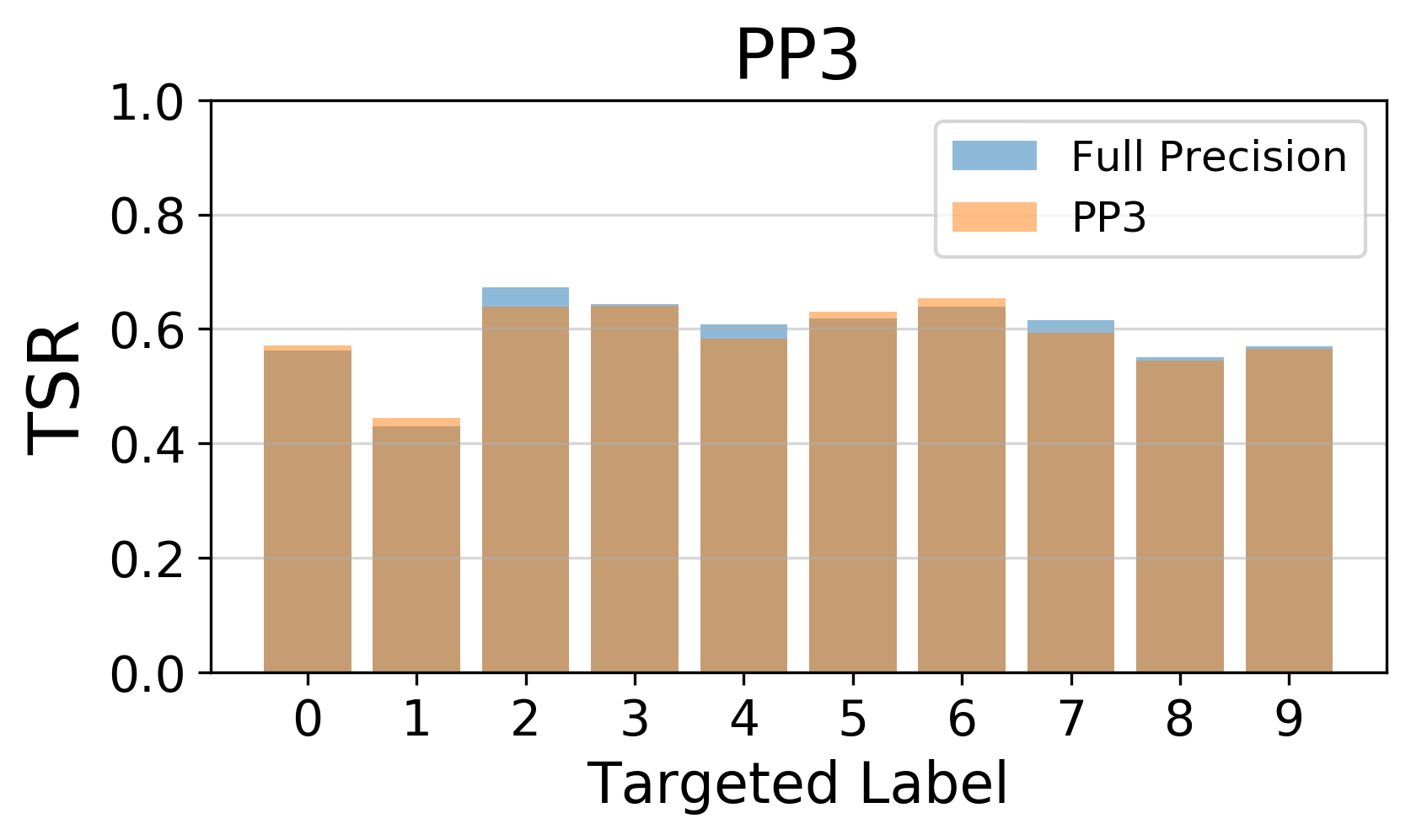}
\includegraphics[width=0.65\columnwidth]{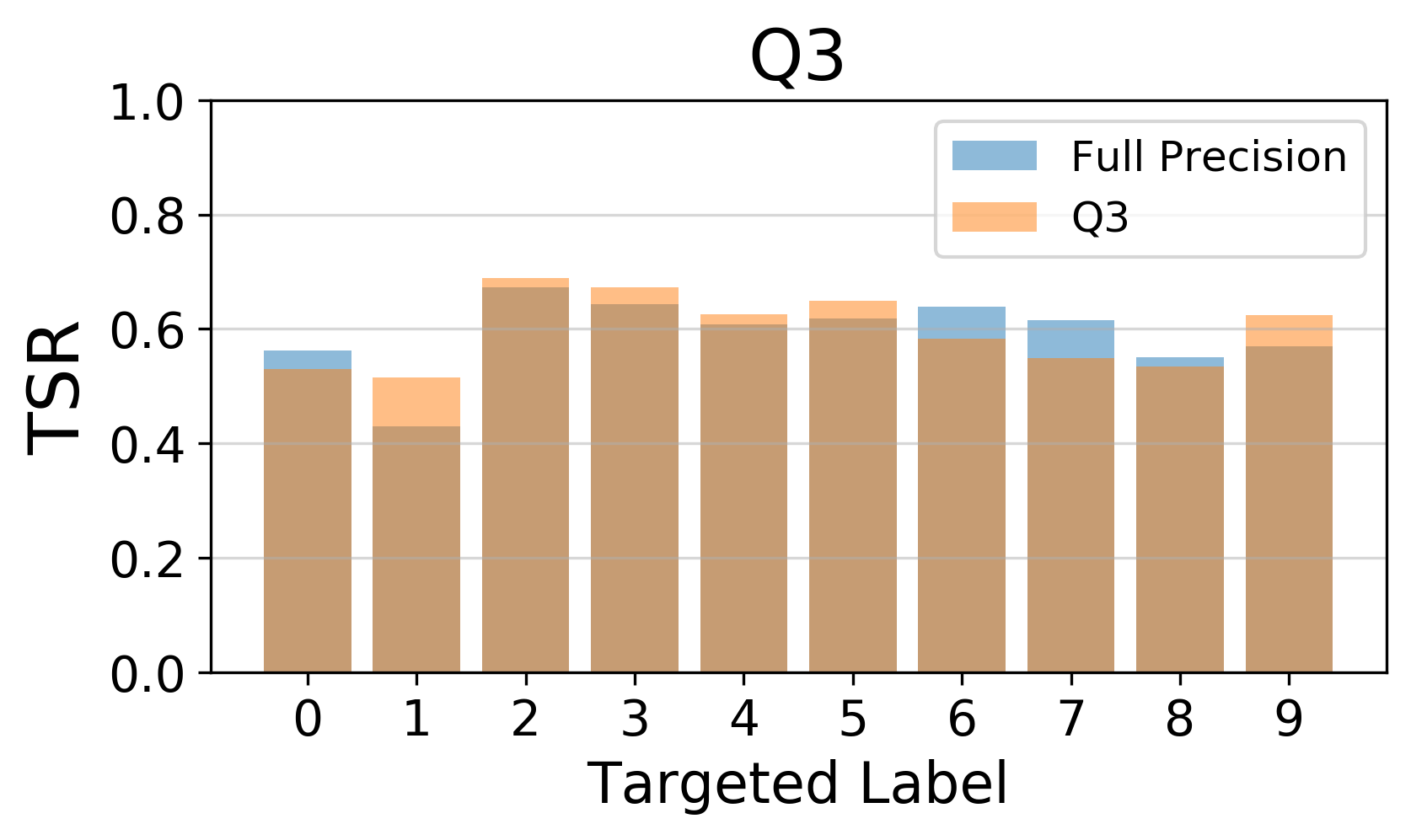}\\
\includegraphics[width=0.65\columnwidth]{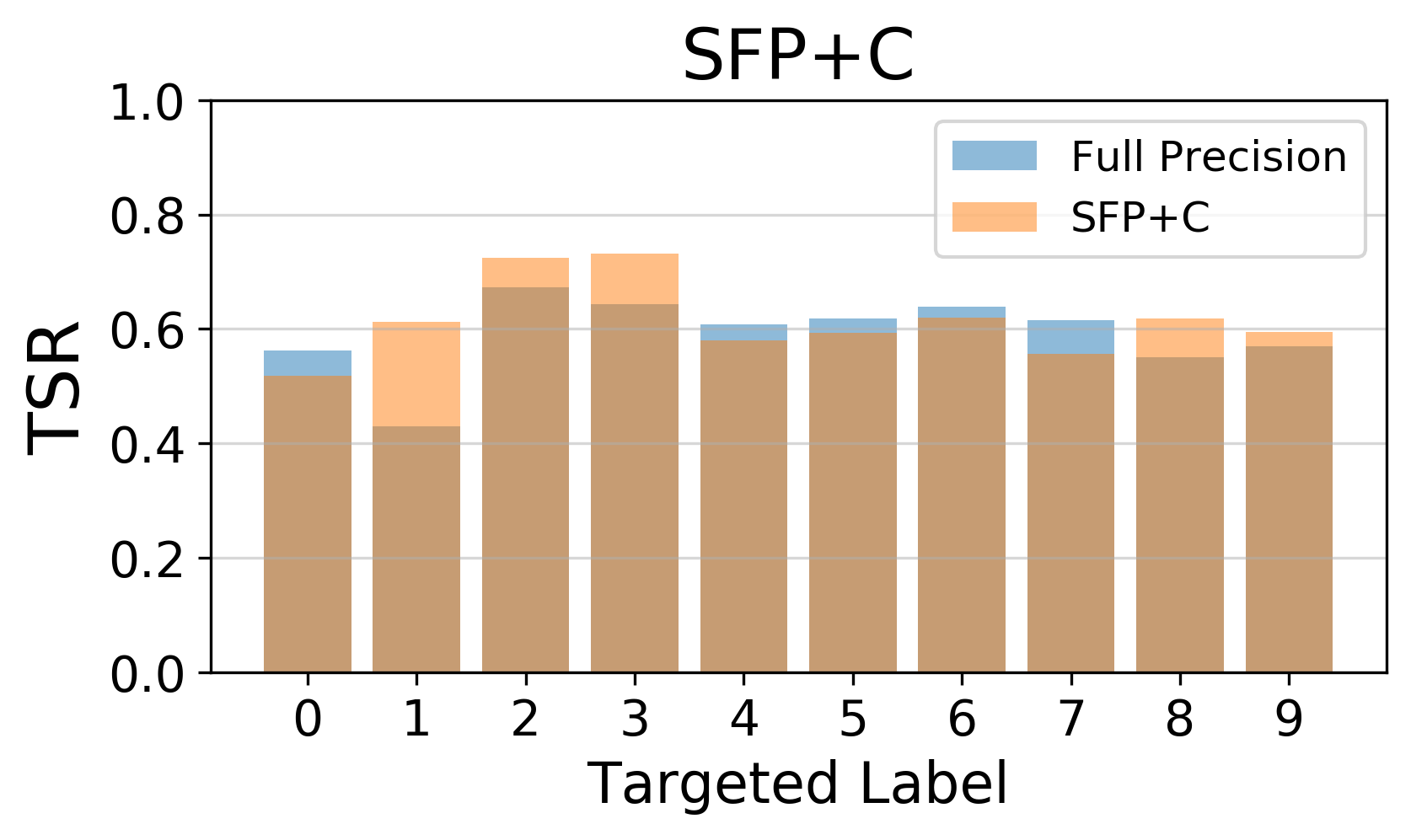}
\includegraphics[width=0.65\columnwidth]{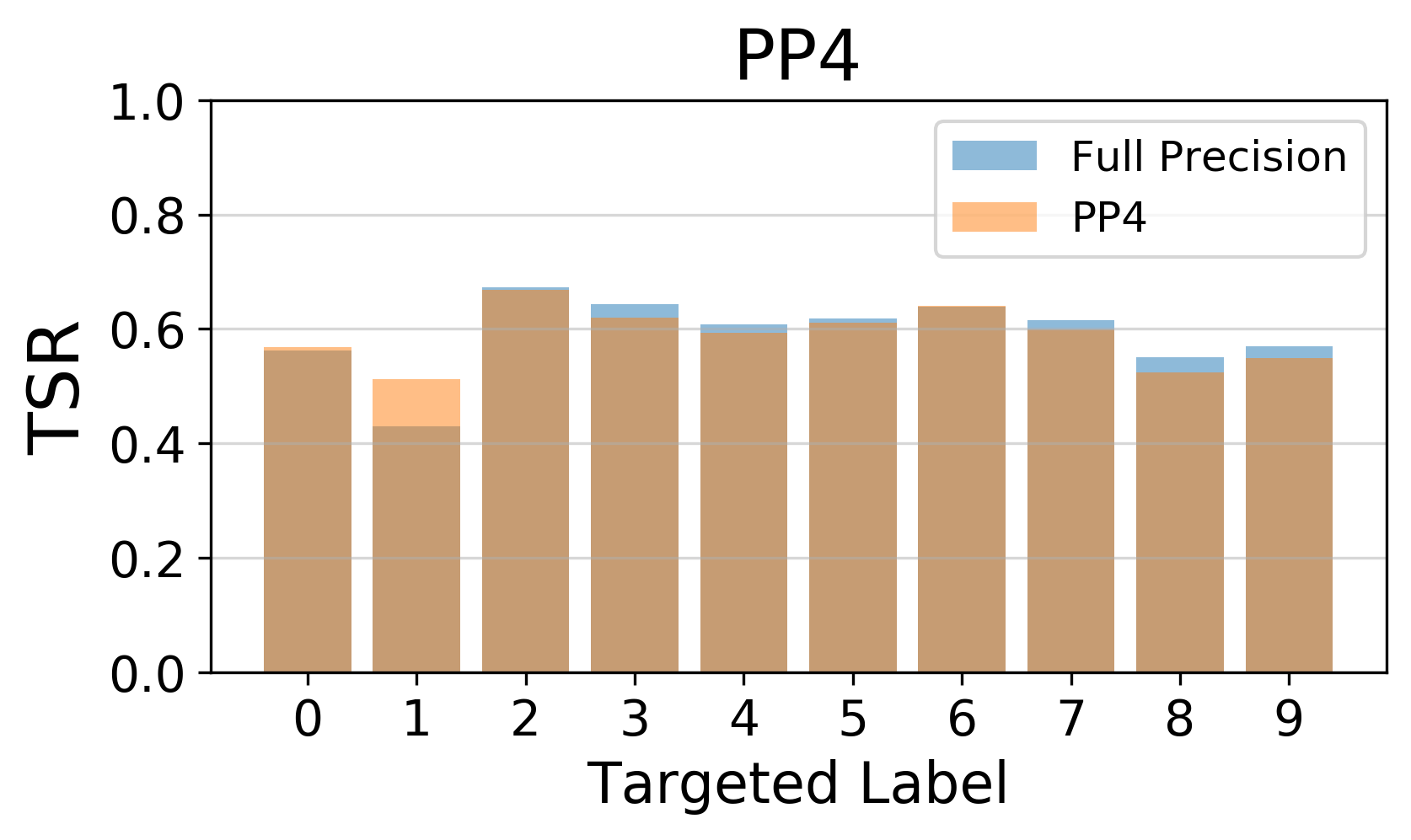}
\includegraphics[width=0.65\columnwidth]{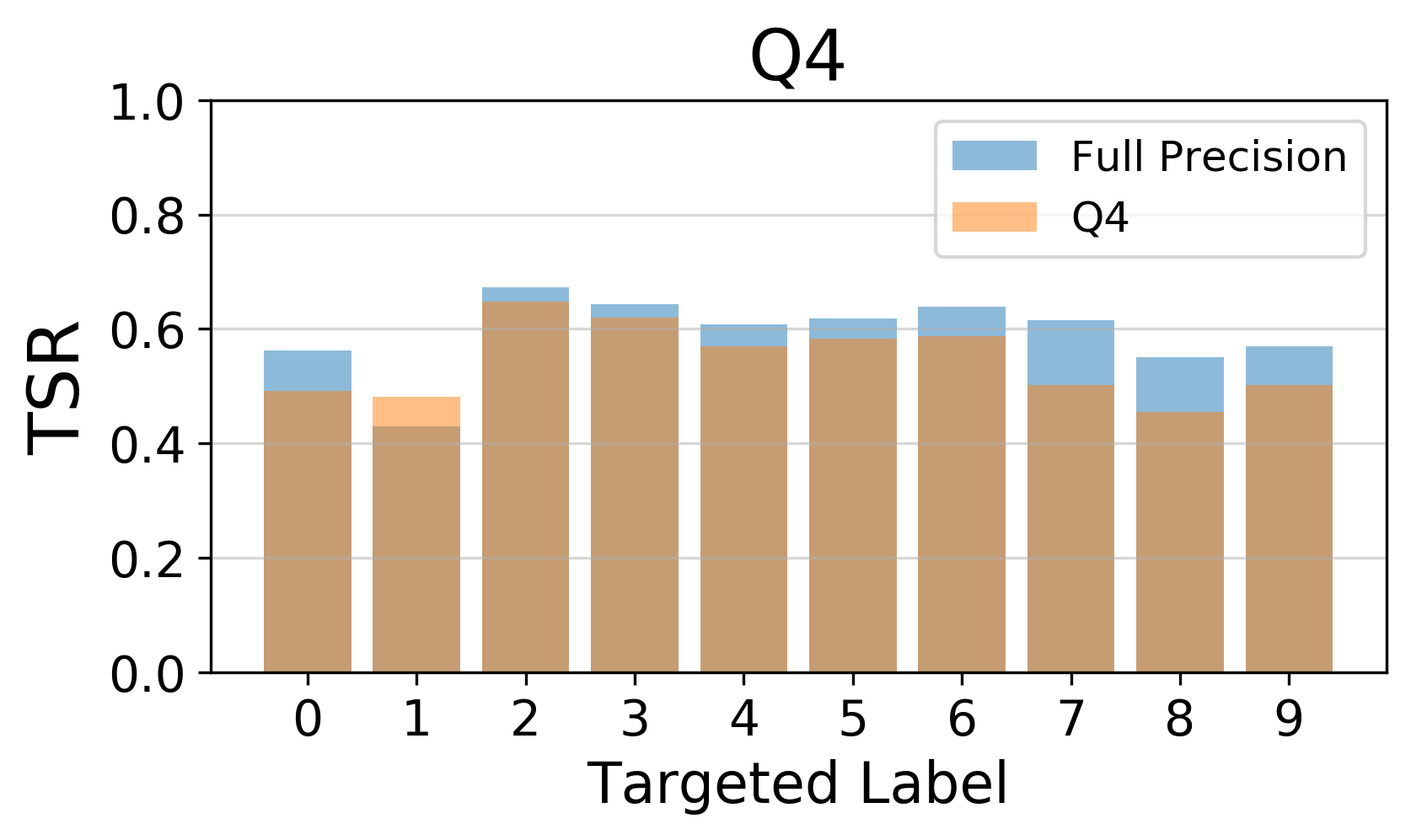}\\
\caption{Targeted UAP results for models on SVHN, with the same template as Figure~\ref{fig:targeted_1}.}
\label{fig:targeted_2}
\end{figure*}

\subsection{UAP Transferability}
We conduct experiments on cross-model adversarial transferability of UAPs at $\varepsilon = 10$. Adversarial samples are generated on the baseline models in Table~\ref{table:clean} (shown on the horizontal axis as ``Attack Source") and tested on the networks displayed vertically as ``Model". Results on both datasets are shown in Figure~\ref{fig:transfer}.

The transferability results reveal that the Full model is mainly vulnerable to the UAPs crafted from the PP$i$ models. Removing neurons post-training yields an adversarial attack as harmful as a white-box attack. However, the adversarial samples crafted from the SFP model are only about half as harmful. This indicates how different the resulting latent space is when pruning during or after training. We note that the Full model's average UER is considerably higher on CIFAR-10 than on SVHN, which is expected given the white-box results discussed previously.

\paragraph{Soft Filter Pruning}
On CIFAR-10 dataset, the SFP model shows noticeably better robustness against UAP transfer attacks from all other models. This model achieved the lowest baseline accuracy, as shown in Table~\ref{table:clean}. Our results suggest that adversarial images are more susceptible to transfer attacks between networks sharing the same or similar baseline accuracies due to their related feature mappings. Furthermore, our results directly contradict those found by \citet{acc_correlated}, in which they hold that adversarial images originating from better performing models are more transferable to less accurate networks. 

We see that SFP models trained on SVHN are more robust against UAP attacks from all other models. When adding regularization to SFP, this robustness becomes even more discernible. Only in the case of SFP plus regularization (mixup and cutout) we observe a lack of transferability to the other models.

We explain this phenomenon by inspecting how each regularization procedure with SFP augments the training data. Mixup adds convex combinations of pairs of examples and their labels to the training data, while cutout adds occluded versions of the inputs. For instance, cutout on CIFAR-10 could partially occlude the object, without necessarily changing the true label of the image, whereas on SVHN it could remove the number, changing the true label of the image. Mixup regularization could have similarly forced the model to use different features when performing classification, as they both still maintain relatively high clean accuracy on SVHN. The differences in the feature sets they learn for the classification task could explain why UAP attacks from other models are not as transferable to SFP with regularization and why UAPs from SFP with regularization are not effective against all the other models trained on SVHN.

\paragraph{Post-training Pruning}
When comparing post-training pruning with the full precision model on CIFAR-10 dataset, we note that the effectiveness of UAPs across all these models is consistent. This indicates that this form of pruning had little effect on the transferability of the generated UAPs. This also appears to be the case when looking at these models on the SVHN dataset. These results suggest that UAP does not exploit neurons or filters that are barely used (activated) for regular inputs but, on the contrary, they exploit combined activations of neurons that are commonly activated for classifying benign inputs. 

\paragraph{Quantization}
On CIFAR-10, we observe that quantization appears to have improved robustness against white-box UAPs, with their UAPs having around 54-63\% UER. However, when considering transfer attacks from other models like the Full or PP$i$ models, these UAPs achieve 67\% and higher UER against the quantized models. Similarly, the effectiveness of transfer attacks from quantized models to the Full and pruned models is also limited. Thus, we show that the black-box attack based on transfer from substitute models overcomes the gradient-masking observed in the previous section. This gradient-masking by quantization is in line with observations from \citet{gupta2020improved}.

However, on SVHN, quantization does not seem to greatly change the robustness to UAPs when compared to the full precision models. These results are consistent with Fig.~\ref{fig:whitebox}, where we show that none of the quantized models achieve better robustness to white-box UAPs on SVHN.

% \begin{figure*}[!ht]
% \centering
% \includegraphics[width=0.5\columnwidth]{figures/tgt-resnet18_sfP-cifar10.png}
% \includegraphics[width=0.5\columnwidth]{figures/tgt-resnet18_P2-0.3-cifar10.png}
% \hspace{1em}
% \includegraphics[width=0.5\columnwidth]{figures/tgt-resnet18_sfP-svhn.png}
% \includegraphics[width=0.5\columnwidth]{figures/tgt-resnet18_P2-0.3-svhn.png}\\
% \includegraphics[width=0.5\columnwidth]{figures/tgt-resnet18_sfP-mixup-cifar10.png}
% \includegraphics[width=0.5\columnwidth]{figures/tgt-resnet18_P3-0.3-cifar10.png}
% \hspace{1em}
% \includegraphics[width=0.5\columnwidth]{figures/tgt-resnet18_sfP-mixup-svhn.png}
% \includegraphics[width=0.5\columnwidth]{figures/tgt-resnet18_P3-0.3-svhn.png}\\
% \includegraphics[width=0.5\columnwidth]{figures/tgt-resnet18_sfP-cutout-cifar10.png}
% \includegraphics[width=0.5\columnwidth]{figures/tgt-resnet18_P4-0.3-cifar10.png}
% \hspace{1em}
% \includegraphics[width=0.5\columnwidth]{figures/tgt-resnet18_sfP-cutout-svhn.png}
% \includegraphics[width=0.5\columnwidth]{figures/tgt-resnet18_P4-0.3-svhn.png}
% \caption{Targeted UAP results for SFP models on both datasets.}
% \label{fig:targeted}
% \end{figure*}

\subsection{Targeted UAPs}
The results for targeted UAPs on CIFAR-10 and SVHN are presented in Figures \ref{fig:targeted_1} and \ref{fig:targeted_2}, respectively. For both datasets, we craft targeted UAPs for each of the 10 class labels at $\varepsilon = 10$. The full precision ResNet18 on CIFAR-10 has an average TSR of 55\% in which targeted UAPs for labels 4 (\emph{deer}), 6 (\emph{frog}), 9 (\emph{truck}) have the highest targeted success rate. SFP increases the robustness of the model by reducing the targeted success rate on 7 out of the 10 classes, only labels 3 (\emph{cat}), 5 (\emph{dog}), 8  (\emph{ship}) have suffered an increase in TSR. Adding regularization produces a sharp increase in robustness. Despite mixup or cutout, all the labels have a lower TSR when compared to that of the Full model. In the case of PP, it appears to weaken the model against some targeted UAPs, particularly for labels 2 (\emph{bird}), 3 (\emph{cat}), 8  (\emph{ship}). For quantized models we observe a reduction of the TSR for most target classes compared to the Full model. However, this is likely due to the gradient masking effect as we have explained before. Thus, more effective UAPs against quantized models can be done via transfer attacks.

On SVHN, the results are similar across the compressed models. Experiments on the full precision model show an evenly spread targeted success rate in the range 55-67\% TSR, except for UAP targeting class 1 which has 43\% TSR. When we apply SFP or SFP+C, the model has slightly reduced robustness. However, SFP+M has contrasting results as targeted UAPs for class labels 3 and 8 have drastically reduced TSR. These results are consistent with our analysis of white-box and transfer attacks, showing that for this dataset the differences in robustness between the Full model and the compressed ones are small in most cases.

\section{Discussion}
\label{sec:discussion}

\subsection{Compression Method Performance}
% CHANGE
The robustness of the compression methods to UAP attacks shows interesting differences between SFP and the other compression methods (PP, Q). SFP reduces the transferability between attacks crafted from other compressed or non-compressed models for both datasets. The reason is that, as SFP removes neurons and filters at training time, it changes the learned feature space compared to the other techniques. This reduces the success rate of transfer attacks from the full or other compressed models. PP and quantization do not seem to affect significantly the robustness to UAP attacks when compared to the full precision model, as the structure of the model is not significantly changed. Although we observe some improvements when using quantization against white-box attacks, we have shown that quantization can produce a gradient-masking effect, but the quantized models are still vulnerable to transfer attacks.

\subsection{CIFAR-10 vs SVHN}
Our empirical evaluation on CIFAR-10 and SVHN shows that the application and the properties of the datasets play an important role in the robustness of the considered compression techniques to UAP attacks. For instance, we observe that even if both, CIFAR-10 and SVHN, have low resolution images with the same number of pixels and class labels, the robustness of SFP to UAP attacks on these datasets leads to different conclusions. Additionally, the regularization method, like cutout, had different effects on the datasets as it is more appropriate for CIFAR-10 but not SVHN.

We also observe a correlation between clean model accuracy and UER of untargeted white-box attacks. Models trained on SVHN achieve better accuracy in general compared to those on CIFAR-10, which translates to lower UER for all perturbation levels $\varepsilon$. In this sense, the variety of shapes and colours for the objects present in CIFAR-10 images can favor the success of UAPs compared to SVHN.

% Our preliminary experiments suggest that the robustness-compression trade-offs to UAP attacks require further evaluation on a wider variety of datasets. Our evaluation shows that even by using comparable computer vision benchmarks, partial results on very specific datasets can lead to erroneous conclusions that may not hold in many practical scenarios. 

\subsection{Practical Implications}
Although compression techniques do not grant immunity against UAP attacks, we see variations in the protection that each compression method gives, with a notable difference especially with transfer attacks. Thus, the model owner can at least make an informed decision in their machine learning pipeline on which methods will better protect their models from universal attacks.

In situations where the model has to be deployed in small devices, compression is non-negotiable due to the application requirements. For this case, the model owner can select a compression technique that, on top of the compression, allows them to defend against adversarial attacks. As the attacks are not as easily transferable across the different compression techniques, this could be used as a defense since the model owner can just roll out different compressed versions of the same model with a specific policy. This policy can then be designed from a security perspective to minimize transferability of UAP attacks between models. In practice, reduced transferability of UAPs can be used to help mitigate the effects of universal attacks. A model owner can deploy the same base model but with different compression techniques across several devices so that a single UAP attack will not have the ability to totally compromise all deployed models. This would increase the attacker's costs as they will then have to craft universal attacks for each individual model.

\section{Related Work}
In this section, we compare our approach and results with recent work that have explored the effect of compression techniques on the robustness of DNNs against adversarial attacks \cite{RW1,RW3, RW2, RW5, RW4}. These works have focused on robustness against input-specific adversarial examples, whereas, to our best knowledge, the analysis of robustness to UAPs proposed in our paper is novel.

\citet{RW3} applied Stochastic Activation Pruning (SAP) to a pre-trained network, without requiring any additional training, which the authors claimed that adds a degree of robustness to the model when against Fast Gradient Sign Method (FGSM) attacks. However, this work is unsuccessful in performing black-box attacks to corroborate their findings on robustness. Thus, improvements in robustness to white-box attacks could be due to gradient masking. Compared to their study, our work proposes a more complete evaluation strategy including not only white-box, but also transfer attacks. \citet{RW5} designed a Defensive Quantization (DQ) method for translating the model into low-bit representation in a more robust manner than regular quantization. Their results against white-box attacks on CIFAR-10 and SVHN are consistent with ours: Quantization models have better robustness than full precision models on CIFAR-10 and a similar performance for SVHN. Their Normal+DQ results against input-specific black-box attacks are also comparable to our results found in UAP transferability. Both their black-box study and ours agree that, the quantized models suffer from inferior performance than the full model on CIFAR-10, and that the Q and Full models achieve comparable accuracy on SVHN. \citet{RW4} found that adversarial examples remain transferable for both pruned and quantized models when evaluating on input-specific attacks. In our analysis of UAP transferability, this appeared to hold true for the PP and Q models. However, the transferability of UAP attacks with SFP suggest that the systemic vulnerabilities of SFP and the full model are different in both datasets. This is possible because SFP prunes the network during training, changing the features learned by the model w.r.t. the full one. Compared to \cite{RW4} our analysis is more comprehensive and includes multiple pruning methods and analyzes the robustness from the perspective of UAPs rather than input-specific attacks. Overall, we add new insight on the effects of compression techniques to adversarial robustness.

\section{Conclusion}
\label{sec:conclusion}

%%% Version 1
% Our work presents a novel analysis on the effects of various compression methods on the models' robustness to Universal Adversarial Perturbations (UAPs). UAPs can deceive machine learning models on a large set of inputs, which makes them a very dangerous and usable attack in practice. We found that pruning during training in the form of SFP was able to improve robustness more reliably when compared to post-training pruning and quantization.

% We show that the various compression methods, and SFP in particular, have significant effects on the transferability of universal attacks across different compression methods. We highlight that a decreased transferability of universal attacks can be used as a defense advantage. In an edge deployment scenario, deploying different compressions in different devices will force the attacker to build particular adversarial samples for each device. The attacker will have larger costs for their attack since the compressed models do not have as much overlap in their vulnerabilities to UAPs.

% Finally, our work shows that various compression techniques have noticeably different results depending on the dataset. This is an important finding as the effect of the compression with respect to the task has to be considered when formulating conclusions on the effect of compression to overall robustness. Our work provides a blueprint for evaluating UAP attacks on various compression techniques, and we hope that future work can expand this evaluation across more tasks, datasets, and compression methods.

%%% Version 2
In this paper we provide an empirical evaluation of the robustness of different compression methods, including pruning and quantization, to UAP attacks against DNNs. Compared to input-specific attacks, UAP attacks provide a more practical perspective as 1) they analyze the systemic vulnerabilities of the models and, 2) most practical attacks (especially physically-realizable ones) exploit universality as they need to be robust to changes in the inputs. Our results show that, overall, the robustness of compressed models is comparable to their original (uncompressed) counterparts. However, pruning at training time with SFP improves the model's robustness to transfer attacks, which can be used to devise defensive strategies when deploying machine learning models on edge devices. We also observe that quantization can give a false sense of security, producing a gradient masking effect. Finally, we show that performance and robustness to UAPs for these compression techniques depend significantly on the structure and the properties of the datasets.

\section{Acknowledgments}
Kenneth Co is supported in part by the DataSpartan research grant DSRD201801.

\bibliography{aaai21}
\end{document}